\documentclass{article}
\usepackage{iclr2026_conference,times}

\usepackage{amsmath,amsfonts,bm}

\def\eqref#1{equation~\ref{#1}}

\def\1{\bm{1}}

\DeclareMathAlphabet{\mathsfit}{\encodingdefault}{\sfdefault}{m}{sl}
\SetMathAlphabet{\mathsfit}{bold}{\encodingdefault}{\sfdefault}{bx}{n}

\usepackage{amsfonts}
\usepackage{amsmath, amsthm, amssymb}

\usepackage[table]{xcolor}
\usepackage{colortbl}
\usepackage{booktabs}
\usepackage{tabularx}
\usepackage{multirow}
\usepackage{makecell}
\usepackage{subcaption}
\usepackage{graphicx}
\usepackage{adjustbox}

\usepackage{appendix}
\usepackage{etoc}
\usepackage{url}
\usepackage{minitoc}

\usepackage{algorithm}
\usepackage{algpseudocode}

\usepackage{listings}
\usepackage{courier}
\usepackage[hidelinks]{hyperref}

\definecolor{paperblue}{RGB}{222,235,247}

\title{WaterSearch: Exploring Seed Pooling for Improving the Quality–Detectability Trade-off in LLM Watermarking}

\author{Yukang Lin$^{1}$, Jiahao Shao$^{1}$, Shuoran Jiang$^{1}$, Wentao Zhu$^{1}$, Bingjie Lu$^{1}$, Xiangping Wu$^{1}$, \\ \textbf{Joanna Siebert}$^{1}$, \textbf{Qingcai Chen}$^{1,2}$\thanks{Corresponding author: qingcai.chen@hit.edu.cn} \\
        $^{1}$Harbin Institute of Technology, Shenzhen, 
        $^{2}$Peng Cheng Laboratory \\
}

\iclrfinalcopy 
\begin{document}


\maketitle

\begin{abstract}
Watermark safeguards the accountability and trust of text generated by Large Language Models (LLMs). Existing watermarking methods typically rely on token-level probability perturbations, which introduce a well-known trade-off between detectability and text quality. In this work, we revisit this problem from the seed selection perspective of the watermark embedding process and observe that watermark behavior varies across random seeds, leading to different trade-offs between detection strength and generation quality. Motivated by this insight, we propose seed-level control via a seed pool to generate diverse candidates and select outputs that best balance quality and detectability. We instantiate this idea in a practical framework, WaterSearch, which enables efficient candidate generation and selection while remaining compatible with existing watermarking methods. Experiments across multiple tasks and model settings demonstrate consistent improvements over conventional watermarking methods, particularly in challenging scenarios including short-text and low-entropy generation. Furthermore, our method is robust against various watermark attacks, including paraphrasing and token-level perturbations. Our code is available at \href{https://github.com/Yukang-Lin/WaterSearch}{https://github.com/Yukang-Lin/WaterSearch}.

\end{abstract}

\section{Introduction}

Large language models (LLMs) have demonstrated strong capabilities in generating high-quality content across diverse domains~\citep{qwen3}. However, their increasingly human-like outputs raise concerns over misuse, including fake news propagation~\citep{misuse1}, harmful content generation~\citep{misuse2}, and copyright infringement~\citep{copyright}. These challenges highlight the need for reliable mechanisms to identify machine-generated content. Watermarking has emerged as a practical technique by embedding statistical signals into generated text, enabling efficient and robust detection~\citep{llm-survey}. A common strategy is to perturb the token-level probability distribution during generation; for example, the KGW framework biases the probability mass of a subset of the vocabulary~\citep{kgw,kgw-v2,gpt-wm}. While such methods often preserve surface-level fluency, prior work has shown that they may degrade downstream performance in scenarios such as short-text generation, reasoning, and low-entropy generation~\citep{downstream}. Recent efforts attempt to mitigate this issue by adapting watermark strength through entropy-aware heuristics or unbiased probability adjustments~\citep{sweet,unbiased}. However, these approaches still rely on modifying the token distributions during generation, which can introduce deviations from the model’s original behavior. As a result, balancing watermark detectability and generation quality remains a practical challenge.

\begin{figure}
    \centering
    \includegraphics[width=\linewidth]{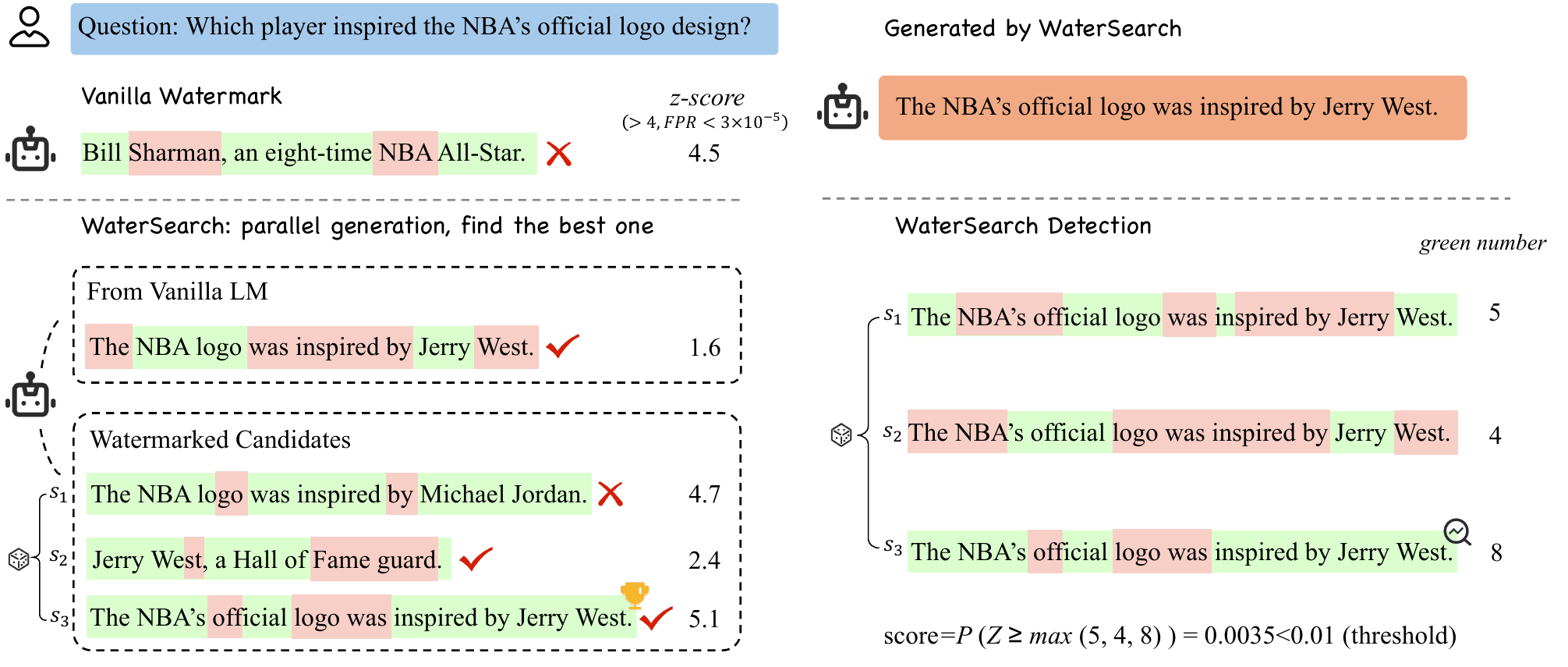}    
    \caption{The mechanism and advantage of WaterSearch. Though the baseline successfully injects a watermark, it compromises text truthfulness (Top Left). In contrast, WaterSearch first choose three seeds and then generate one standard and three watermarked predictions. After that, the best response is selected by a weighted selection metric to balance generation quality and statistical detectability (Bottom Left). For detection, seeds are recover to compute the watermark probability (Right).}
    \label{fig:intro}
\end{figure}

To address this challenge, we revisit watermarking from a different perspective. Instead of focusing solely on token-level perturbations, we consider the role of randomness in watermark generation. In particular, we notice that watermark behavior varies across various random seeds, yielding diverse trade-offs between detection strength and generation quality~\citep{watme}. Based on that, we shift from single-seed generation to seed-level control. Rather than relying on a fixed seed, we maintain a pool of seeds during generation to explore diverse watermark behaviors, producing multiple candidate sequences with different quality–detectability trade-offs. Furthermore, we note that watermark detection is inherently tied to the generation seed, which introduces a form of statistical bias that can be exploited for more effective detection, as illustrated in Figure \ref{fig:kde_freqs}.

Based on these insights, we propose WaterSearch, a sentence-level framework that leverages seed pooling to improve watermarking performance while remaining compatible with a wide range of existing watermarking methods. During generation, WaterSearch produces candidates under different seeds and selects outputs that balance fidelity to the unwatermarked model (quality) and statistical watermark strength (detectability). To make this process efficient, we employ parallel generation together with Key-Value cache reuse. We further provide an intuitive analysis of how seed-level variation affects watermark behavior and its impact on the quality–detectability trade-off. In detection, WaterSearch performs chunk-level hypothesis testing based on a $\chi^2$ statistic, which is robust to token-level perturbations and paraphrasing attacks. Extensive experiments demonstrate that WaterSearch consistently improves over strong baselines across diverse tasks, with particularly notable gains in challenging scenarios such as short-text and low-entropy generation, highlighting the effectiveness of seed-level control in practical watermarking settings.

In summary, our contributions are threefold:
\begin{itemize}
\item We study the effect of random seeds on watermark behavior and introduce seed pooling as a new perspective to improve the watermark quality–detectability trade-off.
\item We propose WaterSearch, a practical framework that generates and selects candidates under different seeds, enabling better control of watermark quality and detectability while remaining compatible with existing methods.
\item Extensive experiments demonstrate consistent improvements across tasks, with notable gains in low-entropy (47.8\%) and short-text generation (36.5\%), as well as robustness to various watermark attacks.
\end{itemize}

\section{Related Work}
\subsection{Watermarking Methods in LLMs}
Text watermarking embeds imperceptible statistical signals into generated text for reliable machine-generated content identification~\citep{christ}. In LLMs, watermarking is typically implemented by modifying token-level probability distributions during generation. The KGW framework~\citep{kgw} partitions the vocabulary and biases the sampling distribution to encode watermark signals. Subsequent work extends this paradigm from multiple perspectives, including probability invariance~\citep{aar}, improved detectability in low-entropy scenarios~\citep{ewd}, and robustness against attacks~\citep{robust-distortion}. Despite these advances, most methods rely on token-level perturbations under a fixed generation pipeline, without explicitly considering alternative sources of variation.

\subsection{The Detectability-Quality Trade-off}

Balancing detection reliability and text quality is a central challenge in LLM watermarking. While early perplexity-based evaluations suggest minimal impact on fluency~\citep{kgw}, recent studies leveraging LLM-as-a-judge and downstream benchmarks reveal noticeable performance degradation in more challenging scenarios~\citep{waterbench}. In particular, prior work reports performance drops in tasks such as short-text generation~\citep{downstream}, reasoning~\citep{watme}, and low-entropy generation~\citep{sweet}. Recent approaches attempt to mitigate this issue through expectation-consistent sampling~\citep{unbiased} and adaptive watermark strength control~\citep{morphmark}. However, these methods can still introduce discrepancies from the model’s original behavior, making it challenging in practice to balance watermark detectability and generation quality.

\subsection{Generation Strategies in Watermark}
Beyond token-level perturbations, recent work explores improving watermarking through alternative generation strategies. For example, \citet{remark-llm} employ beam search for post-hoc editing, while WaterMax~\citep{watermax} selects high-information chunks during generation. Semantic watermarking methods such as SemStamp~\citep{semstamp} and K-SemStamp~\citep{k-semstamp} rely on rejection sampling in embedding space, and SimMark~\citep{simmark} improves performance by similarity of semantic embedding and rejection sampling. These approaches highlight the potential of leveraging generation diversity to improve watermarking quality. However, they primarily operate at the sampling or search procedure level, with limited consideration of how randomness (e.g., random seeds) shapes watermark behavior. In contrast, our work focuses on seed-level control, treating the random seed as a controllable variable to systematically explore diverse watermark behaviors and improve the quality–detectability trade-off.

\section{Problem Definition}
\subsection{Text Generation Process of LLMs}
Large Language Models (LLMs) generate text in an autoregressive manner by sampling tokens from a conditional probability distribution over a predefined vocabulary $\mathcal{V}$. Given an input prompt $\mathbf{x} = \{x_0, \dots, x_{N-1}\}$, the model generates an output sequence $\mathbf{y} = \{y_0, \dots, y_{T-1}\}$ of $T$ tokens autoregressively.

At each decoding step $t$, the model defines a probability distribution over the next token:
\begin{equation}
\mathbb{P}(y_t \mid \mathbf{x}, \mathbf{y}_{[:t-1]}) 
= \text{softmax}\big(\ell^{(t)}(\mathbf{x}, \mathbf{y}_{[:t-1]})\big),
\end{equation}
where $\ell^{(t)}(\cdot) \in \mathbb{R}^{|\mathcal{V}|}$ denotes the logits produced by the model.

\subsection{Watermarked Text Generation}

Watermark injection aims to embed a detectable pattern into generated text by modifying the probability distribution produced by LLMs. We follow the mainstream KGW framework~\citep{kgw} and explicitly formalize the process as follows: 

Let $\text{key}$ denote a secret key shared between the generation and detection procedures. At each decoding step $t$, a token-level seed $s_t$ is deterministically generated based on the previously generated tokens:
\begin{equation}
s_t = \text{Hash}(\mathbf{y}_{[:t-1]}, \text{key}),
\label{eq:hash_func}
\end{equation}
where the hash function ensures consistency between watermark embedding and detection.

The seed $s_t$ is then used to partition the vocabulary $\mathcal{V}$ into two disjoint subsets: a green list $G_t$ and a red list $R_t = \mathcal{V} \setminus G_t$. The watermark is embedded by biasing the logits of tokens in the green list:
\begin{equation}
\ell_k^{(t)}=
\begin{cases}
\ell_k^{(t)}+\delta, & \text{if } k \in G_t \\
\ell_k^{(t)}, & \text{if } k \in R_t,
\end{cases}
\label{eq:kgw-embed}
\end{equation}
where $\gamma$ controls the relative size of the green list and $\delta$ determines the magnitude of the logit bias.

Under this formulation, the watermarking behavior is jointly determined by $(\gamma, \delta)$ and the sequence of token-level seeds $\{s_t\}$ at each steps. While prior work primarily focuses on tuning $(\gamma, \delta)$, this perspective highlights that the seed sequence itself also plays an important role in shaping both watermark detectability and generation quality.

\subsection{Watermarked Text Detection}
For detection, the process involves statistically analyzing the generated text $\mathbf{y}$ by counting green list tokens ($|\mathbf{y}|_G$) and evaluating against the null hypothesis $H_0$ that the text was produced without following the green list rule. The detection confidence is computed via the z-statistic as follows,

\begin{equation}
    z_{\mathbf{y}}=(|\mathbf{y}|_G-\gamma T)\big/(\sqrt{\gamma(1-\gamma)T}),
    \label{eq:zscore}
\end{equation}
where $T$ represents the total number of generated tokens. When the computed z-score exceeds a predetermined threshold, $H_0$ is rejected, confirming the presence of the watermark.

\begin{figure}[htbp]
    \centering
    \includegraphics[width=0.7\linewidth]{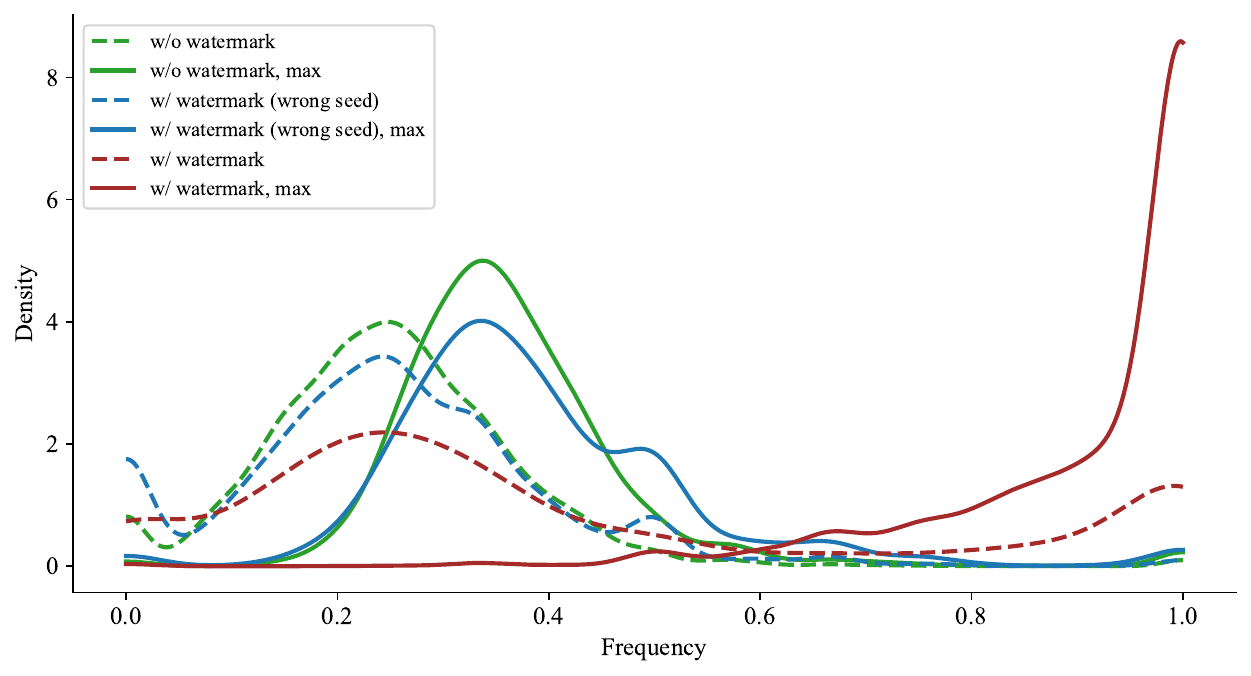}
    \caption{Frequency of token number in green list varying in different seeds under WaterSearch settings. Text without watermark shares a similar distribution with watermarked text in wrong seeds. While under correct seed shows a totally different distribution.}
    \label{fig:kde_freqs}
\end{figure}

\section{WaterSearch: A Search-based Watermarking Framework}
\label{sec:method}
\subsection{Motivation: Candidate Selection}
Under watermarking, generation quality involves two aspects: fidelity to the unwatermarked model and watermark detectability. We formulate this as a candidate selection problem with a sentence-level objective:
{\setlength{\abovedisplayskip}{0pt}
\begin{equation}
    \mathbf{q}(y,\tilde{y}) = \alpha \mathbf{q}_{\text{sim}}(y,\tilde{y}) + (1-\alpha)\frac{|\tilde{y}|_G}{|\tilde{y}|}.
    \label{eq:weighted}
\end{equation}
}

However, standard decoding produces only a single trajectory, limiting the exploration of this trade-off. As watermark behavior depends on token-level seeds, different seeds induce diverse outputs. We therefore introduce seed-level control to generate multiple candidates under different seeds, enabling effective exploration of the quality--detectability space. This forms the basis of WaterSearch.

\subsection{WaterSearch Generation}
Building upon the above formulation, we propose WaterSearch, a search-based watermarking framework that performs chunk-wise parallel generation with seed-level control. The overall procedure is summarized in Algorithm~\ref{algo:WaterSearch-generation}.

\paragraph{Seed Pool Construction.}
To enable diverse watermark behaviors while ensuring reproducibility for detection, we introduce a seed pool mechanism that maintains a set of global seeds at each generation step. Specifically, we construct the seed pool via a key-conditioned permutation:
\begin{equation}
\text{SeedPool} = \pi_{\text{key}}([s_1, s_2, \cdots]),
\label{eq:seed_pool}
\end{equation}
where $[s_1, s_2, \cdots]$ is a predefined seed set, and $\pi_{\text{key}}(\cdot)$ denotes a permutation function parameterized by a secret key, ensuring consistency between generation and detection. Given the key, the detector can reconstruct the ordering of seeds during decoding. In practice, this permutation can be instantiated using a simple shuffle operation conditioned on the key. For each candidate, the token-level seed at each decoding step follows Eq.~(\ref{eq:hash_func}), where the seed is instantiated as $s_i \in \text{SeedPool}$. This design enables controlled variation in watermark behavior while remaining compatible with existing token-level watermarking methods.

\begin{algorithm}[t]
\small
\caption{WaterSearch Generation}
\label{algo:WaterSearch-generation}
\begin{algorithmic}[1]
\State \textbf{Input:} prompt $\mathbf{c}$, chunk size $m$, beam size $k$
\State \hspace{1.35cm} base processor $\ell$, watermark processor $\tilde{\ell}(\cdot)$
\For{$i = 1$ to $\lceil T/m \rceil$}
    \State \textbf{// Seed Sampling}
    \State $[s_1,\dots,s_{k-1}] \gets \text{SeedPool}(\mathbf{c})$
    \Statex\Comment{Sample seeds using Eq.~(\ref{eq:seed_pool})}
    \State \textbf{// Processor Construction}
    \State $\mathcal{L} \gets [\,\ell,~\tilde{\ell}(s_1),\dots,\tilde{\ell}(s_{k-1})\,]$
    \Statex\Comment{Apply watermarking via Eq.~(\ref{eq:kgw-embed})}
    \State \textbf{// Parallel Generation}
    \State $[\mathbf{y}_i^{(0)},\dots,\mathbf{y}_i^{(k-1)}] 
    \gets \text{ParallelGenerate}(\mathbf{c}, \mathcal{L})$
    \Statex\Comment{Generate $k$ candidates}
    \State \textbf{// Candidate Selection}
    \State $\hat{\mathbf{y}}_i \gets 
    \arg\max_{\tilde{\mathbf{y}}} \mathbf{q}(y,\tilde{y})$
    \Comment{Select via Eq.~(\ref{eq:weighted})}
    \State \textbf{// Context Update}
    \State $\mathbf{c} \gets [\mathbf{c} \,\|\, \hat{\mathbf{y}}_i]$
    \Comment{Append selected chunk}
\EndFor
\State \textbf{Output:} $\mathbf{y} = [\hat{\mathbf{y}}_1,\dots,\hat{\mathbf{y}}_{\lceil T/m\rceil}]$
\end{algorithmic}
\end{algorithm}

\begin{algorithm}[t]
\small
\caption{WaterSearch Detection}
\label{algo:WaterSearch-detect}
\begin{algorithmic}[1]
\State \textbf{Input:} prompt $\mathbf{c}$, response $\mathbf{y}$
\State \hspace{1.35cm} chunk size $m$, beam size $k$
\State $\mathcal{P} \gets [\,]$ \Comment{List of p-values}
\State $[\mathbf{y}_1,\dots,\mathbf{y}_{\lceil T/m\rceil}] \gets \text{Split}(\mathbf{y}, m)$
\For{$i = 1$ to $\lceil T/m\rceil$}
    \State \textbf{// Seed Reconstruction}
    \State $[s_1,\dots,s_{k-1}] \gets \text{SeedPool}(\mathbf{c})$ 
    \Comment{Eq.~(\ref{eq:seed_pool})}
    \State \textbf{// Chunk Test}
    \State $p_i \gets \text{ChunkTest}(\mathbf{y}_i, \{s_j\})$ 
    \Comment{Eq.~(\ref{eq:bio-max-distribution}), (\ref{eq:p-val})}
    \State $\mathcal{P}.\text{append}(p_i)$
\EndFor
\State \textbf{// Aggregation}
\State $p_{\text{doc}} \gets \text{Combine}(\mathcal{P})$ 
\Comment{Eq.~(\ref{eq:fisher_combined})}
\State \textbf{Output:} $p_{\text{doc}}$
\end{algorithmic}
\end{algorithm}

\paragraph{Parallel Generation and Selection.}
Given an input prompt $\mathbf{c}$, WaterSearch generates $k$ candidate sequences in parallel at each step, including one unwatermarked text and $k-1$ watermarked candidates under different seeds. Each candidate is generated using a corresponding logit processor, where watermark processors follow Eq.~(\ref{eq:kgw-embed}).

The candidates are then evaluated using Eq.~(\ref{eq:weighted}), and the best candidate is selected:
\begin{equation}
\hat{\mathbf{y}} = \arg\max_{\tilde{\mathbf{y}}} \mathbf{q}(y, \tilde{\mathbf{y}}).
\label{eq:choose_best}
\end{equation}

The selected output is appended to the context, and the process repeats for the next chunk. To improve efficiency, we reuse the KV-cache across candidates during parallel generation.

This chunk-wise search process enables effective exploration of diverse watermarking behaviors while maintaining low computational overhead.

\subsection{WaterSearch Detection}

Detection in WaterSearch is challenging because the exact seed used for generating each selected chunk is not directly observable (choose by Eq.~\ref{eq:choose_best}). Instead of identifying the true seed, we exploit a key statistical property: the correct seed tends to produce higher green-token counts, as supported in Fig.~\ref{fig:kde_freqs}. This observation motivates a max-statistic based detection strategy. Specifically, for each chunk $\mathbf{y}_i$, we evaluate watermark statistics across a set of candidate seeds and consider the maximum response. We formalize the detection problem as a hypothesis test:

\textit{$H_0$: The text is generated without WaterSearch (no seed-level manipulation).} 
\\
\textit{$H_1$: The text is generated by WaterSearch.}

Under $H_0$, for a given seed $s$, the number of green tokens in chunk $\mathbf{y}_i$ is expressed by $z_i(s) = |\mathbf{y}_i|_{G(s)}$, which follows a binomial distribution:

\begin{equation}
z_i(s) \sim \mathcal{B}(|\mathbf{y}_i|, \gamma),
\end{equation}
where $|\mathbf{y}_i|$ is the chunk length and $\gamma$ is the green-list ratio.

We then define the chunk-level test statistic as the maximum over a set of $k-1$ candidate seeds:
\begin{equation}
Z_i = \max_{1 \le j \le k-1} z_i(s_j).
\end{equation}

Under $H_0$, assuming independence across seeds, $Z_i$ follows the maximum of i.i.d. binomial random variables. Its cumulative distribution function is given by Eq.~(\ref{eq:bio-max-distribution}), where $n = |\mathbf{y}_i|$.

\begin{equation}
F(z) = P(Z_i \le z) = \left[ \sum_{t=0}^{z} \binom{n}{t} \gamma^t (1-\gamma)^{n-t} \right]^{k-1}
\label{eq:bio-max-distribution}
\end{equation}

The corresponding p-value for the observed statistic $z_{\text{obs}}$ is:
\begin{equation}
p_i = P(Z_i \ge z_{\text{obs}}) = 1 - F(z_{\text{obs}} - 1).
\label{eq:p-val}
\end{equation}

Finally, we aggregate chunk-level p-values $\{p_i\}$ using Fisher's method:
\begin{equation}
-2 \sum_{i=1}^{\lceil T/m\rceil} \ln p_i \sim \chi^2\big(2\lceil T/m\rceil\big),
\label{eq:fisher_combined}
\end{equation}
which provides a document-level significance score. This detection scheme is inherently robust, as it integrates evidence across multiple chunks, allowing reliable decisions at the document level even when individual chunks provide limited signals. The detailed process is shown in Algorithm~\ref{algo:WaterSearch-detect}.

\section{Experiments}
\subsection{Experimental Setup}
\label{ssec:exp-setup}
We choose WaterBench~\citep{waterbench} as our evaluation benchmark. WaterBench is a multi-task benchmark spanning nine typical NLP tasks with varying input / output lengths. Moreover, it requires a consistent watermarking strength for a fair comparison. We also add the RepoBench-P~\citep{repobench} dataset to supplement the dataset for code scenarios. All the tasks are evaluated with the proper metrics as shown in Table \ref{tab:data_intro}.

We evaluate on three representative LLMs: Qwen2.5-7B-Instruct~\citep{qwen3}, Llama2-7B-Chat~\citep{llama2} and InternLM-7B-Chat~\citep{internlm}. Four representative watermarks are included: KGW-Hard~\citep{kgw}, KGW-Soft~\citep{kgw}, GPT Watermark~\citep{gpt-wm} and V2 Watermark~\citep{kgw-v2}. During the evaluation, watermark strength (True Positive Rate, TPR) is set to 95\% for pairwise comparison of different downstream task performance. For WaterSearch, we set the number of beams $m$ to 5, consisting of 1 standard text without watermark and 4 watermarked texts with distinct seeds. For detection, we set the confidence threshold for the p-value to 0.01. Details of parameter selection are discussed in Appendix \ref{appendix:params}.

\subsection{Main Results}


\begin{table}[htbp]
  \centering
  \small
  \renewcommand{\arraystretch}{1.2}
  \resizebox{\textwidth}{!}{
    \setlength{\tabcolsep}{4pt}  
    \begin{tabular}{c c *{13}{c}}
      \toprule
      \multirow{2}{*}{} & \multirow{2}{*}{\textbf{Method}}
      & \multicolumn{2}{c}{\textbf{Short In/Short Out}} 
      & \multicolumn{2}{c}{\textbf{Short In/Long Out}} 
      & \multicolumn{3}{c}{\textbf{Long In/Short Out}} 
      & \multicolumn{2}{c}{\textbf{Long In/Long Out}} 
      & \multicolumn{1}{c}{\textbf{Open-Ended}} 
      & \multicolumn{3}{c}{\textbf{Overall}}  \\ 
      \cmidrule(lr){3-4} \cmidrule(lr){5-6} \cmidrule(lr){7-9} \cmidrule(lr){10-11} \cmidrule(lr){12-12} \cmidrule(lr){13-15}
      & & \hspace{6pt}\textbf{KoLA} & \textbf{Copen} & \hspace{8pt}\textbf{ELI5} & \textbf{FiQA} 
      & \hspace{2pt}\textbf{HotpotQA} & \textbf{LCC} & \textbf{RepoBench-P}
      & \hspace{2pt}\textbf{Multinews} & \textbf{QMsum} 
      & \textbf{AlpacaFarm} & \textbf{TP} & \textbf{TN} & \textbf{GM} \\
      \midrule
      \multirow{9}{*}{\rotatebox[origin=c]{90}{Qwen-2.5-7B-Instruct}}
      & Hard Watermark
      & \hspace{6pt}6.2 & 4.5 & \hspace{8pt}20.0 & 17.3
      & 13.6 & 20.3 & 15.4 & 17.2 & 14.2
      & 13.3 & 95.6 & 99.5 & 14.2 \\
      & \cellcolor{paperblue}+ WaterSearch
      & \cellcolor{paperblue}\hspace{6pt}13.3 & \cellcolor{paperblue}37.8 & \cellcolor{paperblue}\hspace{8pt}22.6 & \cellcolor{paperblue}18.7
      & \cellcolor{paperblue}29.3 & \cellcolor{paperblue}22.6 & \cellcolor{paperblue}25.1 & \cellcolor{paperblue}19.7 & \cellcolor{paperblue}18.0
      & \cellcolor{paperblue}27.7 & \cellcolor{paperblue}96.4 & \cellcolor{paperblue}99.7 & \cellcolor{paperblue}\textbf{23.5} \\
      & Soft Watermark
      & \hspace{6pt}3.1 & 41.0 & \hspace{8pt}16.5 & 14.1
      & 36.1 & 15.2 & 17.4 & 12.2 & 14.4
      & 6.4 & 94.8 & 99.7 & 17.6 \\
      & \cellcolor{paperblue}+ WaterSearch
      & \cellcolor{paperblue}\hspace{6pt}9.9 & \cellcolor{paperblue}41.0 & \cellcolor{paperblue}\hspace{8pt}19.1 & \cellcolor{paperblue}15.0
      & \cellcolor{paperblue}45.9 & \cellcolor{paperblue}20.2 & \cellcolor{paperblue}23.6 & \cellcolor{paperblue}13.5 & \cellcolor{paperblue}15.8
      & \cellcolor{paperblue}11.1 & \cellcolor{paperblue}96.4 & \cellcolor{paperblue}99.8 & \cellcolor{paperblue}\textbf{21.5} \\
      & GPT Watermark
      & \hspace{6pt}4.6 & 42.0 & \hspace{8pt}5.6 & 5.5
      & 35.3 & 5.0 & 8.4 & 3.6 & 11.7
      & 4.4 & 95.3 & 99.7 & 12.6 \\
      & \cellcolor{paperblue}+ WaterSearch
      & \cellcolor{paperblue}\hspace{6pt}7.9 & \cellcolor{paperblue}40.8 & \cellcolor{paperblue}\hspace{8pt}14.9 & \cellcolor{paperblue}13.0
      & \cellcolor{paperblue}40.0 & \cellcolor{paperblue}17.8 & \cellcolor{paperblue}16.8 & \cellcolor{paperblue}10.8 & \cellcolor{paperblue}15.2
      & \cellcolor{paperblue}7.1 & \cellcolor{paperblue}97.3 & \cellcolor{paperblue}99.1 & \cellcolor{paperblue}\textbf{18.4} \\
      & V2 Watermark
      & \hspace{6pt}5.1 & 23.3 & \hspace{8pt}19.3 & 15.5
      & 16.6 & 12.5 & 13.5 & 13.5 & 13.7
      & 7.1 & 95.5 & 99.8 & 14.0 \\
      & \cellcolor{paperblue}+ WaterSearch
      & \cellcolor{paperblue}\hspace{6pt}5.4 & \cellcolor{paperblue}38.0 & \cellcolor{paperblue}\hspace{8pt}21.7 & \cellcolor{paperblue}17.7
      & \cellcolor{paperblue}32.0 & \cellcolor{paperblue}22.0 & \cellcolor{paperblue}22.2 & \cellcolor{paperblue}17.2 & \cellcolor{paperblue}16.3
      & \cellcolor{paperblue}20.6 & \cellcolor{paperblue}98.9 & \cellcolor{paperblue}99.7 & \cellcolor{paperblue}\textbf{21.3} \\
      \hline

      \multirow{9}{*}{\rotatebox[origin=c]{90}{Llama2-7B-Chat}}
      & Hard Watermark
      & \hspace{6pt}1.1 & 8.9 & \hspace{8pt}10.5 & 13.6
      & 4.9 & 27.8 & 25.9 & 11.1 & 12.2
      & 1.1 & 95.6 & 99.5 & 11.7 \\
      & \cellcolor{paperblue}+ WaterSearch
      & \cellcolor{paperblue}\hspace{6pt}2.4 & \cellcolor{paperblue}28.0 & \cellcolor{paperblue}\hspace{8pt}13.7 & \cellcolor{paperblue}16.4
      & \cellcolor{paperblue}8.6 & \cellcolor{paperblue}29.0 & \cellcolor{paperblue}27.2 & \cellcolor{paperblue}15.0 & \cellcolor{paperblue}14.0 
      & \cellcolor{paperblue}4.5 & \cellcolor{paperblue}98.7 & \cellcolor{paperblue}99.5 & \cellcolor{paperblue}\textbf{15.9} \\
      & Soft Watermark
      & \hspace{6pt}1.7 & 13.8 & \hspace{8pt}8.1 & 11.8
      & 14.4 & 25.3 & 22.2 & 9.3 & 11.0
      & 0.6 & 95.3 & 99.5 & 11.8 \\
      & \cellcolor{paperblue}+ WaterSearch
      & \cellcolor{paperblue}\hspace{6pt}2.4 & \cellcolor{paperblue}27.2 & \cellcolor{paperblue}\hspace{8pt}12.4 & \cellcolor{paperblue}16.0
      & \cellcolor{paperblue}16.3 & \cellcolor{paperblue}27.6 & \cellcolor{paperblue}27.1 & \cellcolor{paperblue}13.6 & \cellcolor{paperblue}13.5
      & \cellcolor{paperblue}4.5 & \cellcolor{paperblue}97.6 & \cellcolor{paperblue}99.6 & \cellcolor{paperblue}\textbf{16.1} \\
      & GPT Watermark
      & \hspace{6pt}1.8 & 25.3 & \hspace{8pt}4.5 & 5.9
      & 12.5 & 17.0 & 12.9 & 4.8 & 9.6
      & 0.2 & 97.0 & 96.9 & 9.5 \\
      & \cellcolor{paperblue}+ WaterSearch
      & \cellcolor{paperblue}\hspace{6pt}1.6 & \cellcolor{paperblue}31.0 & \cellcolor{paperblue}\hspace{8pt}11.1 & \cellcolor{paperblue}13.5
      & \cellcolor{paperblue}17.4 & \cellcolor{paperblue}21.2 & \cellcolor{paperblue}21.9 & \cellcolor{paperblue}11.0 & \cellcolor{paperblue}13.2 
      & \cellcolor{paperblue}4.5 & \cellcolor{paperblue}96.5 & \cellcolor{paperblue}99.0 & \cellcolor{paperblue}\textbf{14.6} \\
      & V2 Watermark
      & \hspace{6pt}1.1 & 21.3 & \hspace{8pt}13.2 & 13.5
      & 7.4 & 20.4 & 23.3 & 11.7 & 11.5 
      & 0.9 & 94.5 & 99.9 & 12.4 \\
      & \cellcolor{paperblue}+ WaterSearch
      & \cellcolor{paperblue}\hspace{6pt}2.3 & \cellcolor{paperblue}33.6 & \cellcolor{paperblue}\hspace{8pt}14.6 & \cellcolor{paperblue}16.8
      & \cellcolor{paperblue}15.2 & \cellcolor{paperblue}27.8 & \cellcolor{paperblue}27.8 & \cellcolor{paperblue}16.2 & \cellcolor{paperblue}13.9 
      & \cellcolor{paperblue}5 & \cellcolor{paperblue}99.4 & \cellcolor{paperblue}99.8 & \cellcolor{paperblue}\textbf{17.3} \\
      \hline

      \multirow{9}{*}{\rotatebox[origin=c]{90}{InternLM-7B-Chat}}
      & Hard Watermark
      & \hspace{6pt}2.8 & 0.8 & \hspace{8pt}10.7 & 8.4
      & 3.2 & 20.1 & 15.8 & 5.3 & 7.4
      & 0.8 & 93.3 & 99.7 & 7.5 \\
      & \cellcolor{paperblue}+ WaterSearch
      & \cellcolor{paperblue}\hspace{6pt}2.6 & \cellcolor{paperblue}6.1 & \cellcolor{paperblue}\hspace{8pt}13.8 & \cellcolor{paperblue}12.9
      & \cellcolor{paperblue}3.5 & \cellcolor{paperblue}20.9 & \cellcolor{paperblue}20.7 & \cellcolor{paperblue}8.4 & \cellcolor{paperblue}10.6 
      & \cellcolor{paperblue}2.0 & \cellcolor{paperblue}99.8 & \cellcolor{paperblue}99.2 & \cellcolor{paperblue}\textbf{10.2} \\
      & Soft Watermark
      & \hspace{6pt}2.4 & 10.1 & \hspace{8pt}9.1 & 6.1
      & 2.5 & 18.6 & 17.7 & 4.0 & 5.3
      & 0.3 & 94.0 & 99.6 & 7.6 \\
      & \cellcolor{paperblue}+ WaterSearch
      & \cellcolor{paperblue}\hspace{6pt}5.0 & \cellcolor{paperblue}24.6 & \cellcolor{paperblue}\hspace{8pt}16.1 & \cellcolor{paperblue}15.5
      & \cellcolor{paperblue}15.4 & \cellcolor{paperblue}15.3 & \cellcolor{paperblue}15.8 & \cellcolor{paperblue}12.6 & \cellcolor{paperblue}11.5 
      & \cellcolor{paperblue}5.3 & \cellcolor{paperblue}97.1 & \cellcolor{paperblue}99.2 & \cellcolor{paperblue}\textbf{13.7} \\
      & GPT Watermark
      & \hspace{6pt}1.9 & 4.5 & \hspace{8pt}8.5 & 7.1
      & 2.4 & 20.5 & 19.4 & 4.2 & 6.2
      & 0.5 & 95.6 & 99.8 & 7.5 \\
      & \cellcolor{paperblue}+ WaterSearch
      & \cellcolor{paperblue}\hspace{6pt}4.2 & \cellcolor{paperblue}11.0 & \cellcolor{paperblue}\hspace{8pt}15.6 & \cellcolor{paperblue}15.2
      & \cellcolor{paperblue}15.6 & \cellcolor{paperblue}25.9 & \cellcolor{paperblue}17.1 & \cellcolor{paperblue}12.4 & \cellcolor{paperblue}12.4 
      & \cellcolor{paperblue}5.3 & \cellcolor{paperblue}98.4 & \cellcolor{paperblue}98.7 & \cellcolor{paperblue}\textbf{13.5} \\
      & V2 Watermark
      & \hspace{6pt}1.3 & 20.6 & \hspace{8pt}9.0 & 6.3
      & 3.4 & 28.1 & 27.3 & 5.3 & 5.6
      & 0.5 & 94.9 & 99.8 & 10.7 \\
      & \cellcolor{paperblue}+ WaterSearch
      & \cellcolor{paperblue}\hspace{6pt}4.7 & \cellcolor{paperblue}33.5 & \cellcolor{paperblue}\hspace{8pt}15.3 & \cellcolor{paperblue}15.4
      & \cellcolor{paperblue}16.7 & \cellcolor{paperblue}28.8 & \cellcolor{paperblue}33.3 & \cellcolor{paperblue}12.5 & \cellcolor{paperblue}11.5
      & \cellcolor{paperblue}4.5 & \cellcolor{paperblue}95.3 & \cellcolor{paperblue}99.6 & \cellcolor{paperblue}\textbf{17.6} \\
      \bottomrule
    \end{tabular}
  }
  \caption{Performance comparison of WaterSearch and counterpart base watermarking methods across 10 benchmark datasets. The final ``Overall" column summarizes both generation quality (GM) and detection accuracy metrics: True Positive rate (TP, correctly judge the text has a watermark) and True Negative rate (TN, correctly judge the text does not have a watermark).}
  \label{tab:main-result}
\end{table}


\begin{table}[!htbp]
  \centering
  \small
  \renewcommand{\arraystretch}{1.2}
  \caption{Detail results in short text and low entropy generation scenarios.}
    \resizebox{\textwidth}{!}{
    \begin{tabular}{llllllllllllllll}
    \toprule
               & \multicolumn{9}{c}{Short Text Generation} & \multicolumn{6}{c}{Low Entropy Generation} \\
               \cmidrule(lr){2-10} \cmidrule(lr){11-16}
               & \multicolumn{3}{c}{KoLA} & \multicolumn{3}{c}{Copen} & \multicolumn{3}{c}{HotpotQA} & \multicolumn{3}{c}{LCC} & \multicolumn{3}{c}{RepoBench-P} \\
               \cmidrule(lr){2-4} \cmidrule(lr){5-7} \cmidrule(lr){8-10} \cmidrule(lr){11-13} \cmidrule(lr){14-16}
               & TP & TN & GM & TP & TN & GM & TP & TN & GM & TP & TN & GM & TP & TN & GM \\ 
    \midrule
    Hard Watermark & 100.0 & 100.0 & 1.1 & 79.0 & 100.0 & 8.9 & 72.0 & 100.0 & 4.9 & 93.0 & 100.0 & 27.8 & 97.5 & 100.0 & 25.9 \\
    WaterSearch ($\alpha$=0.75) & 100.0 & 98.0 & \textbf{2.4} & 96.4 & 100.0 & \textbf{28.0} & 95.5 & 100.0 & \textbf{8.6} & 95.0 & 99.4 & \textbf{29.0} & 92.0 & 98.0 & \textbf{27.2} \\
    \bottomrule
    \end{tabular}
    }
    \label{tab:short-output-low-entropy}
\end{table}

We evaluate WaterSearch against baseline watermarking methods across ten tasks. Table~\ref{tab:main-result} reports the overall performance relative to their corresponding base methods.  Overall, WaterSearch consistently improves downstream task performance across all settings, with an average absolute gain of around 5.5 points. The improvements remain stable across different model families, including Qwen2.5-7B-Instruct (+6.6), LLaMA2-7B-Chat (+4.6), and InternLM-7B-Chat (+5.4), indicating the robustness and general applicability of the proposed seed-level control mechanism.

We further analyze two challenging scenarios where existing watermarking methods often struggle: short-text generation and low-entropy generation.

\paragraph{Short Text Generation}
In short-text generation, the limited number of tokens makes it difficult for conventional methods to accumulate sufficient watermark evidence for reliable detection. As shown in Table~\ref{tab:short-output-low-entropy}, WaterSearch significantly improves detection success rates across datasets such as Copen and HotpotQA. This improvement can be attributed to the ability of WaterSearch to explore multiple candidates and select sequences with stronger watermark signals, thereby facilitating more reliable detection under limited token budgets.

\paragraph{Low-Entropy Generation}
Watermarking in low-entropy scenarios is particularly challenging, as the token distribution is highly concentrated and difficult to perturb without degrading generation quality. Code completion serves as a representative example, where strong structural constraints make tokens highly predictable. As shown in Table~\ref{tab:short-output-low-entropy}, WaterSearch maintains high text quality while improving detectability, achieving an average gain of 4.7 points in such settings. This is achieved by selecting candidates that remain close to the original model outputs while exhibiting stronger watermark signals, effectively balancing the quality--detectability trade-off.

\section{Attacking Robustness}

\begin{figure}[!h]
\centering
\includegraphics[width=\linewidth]{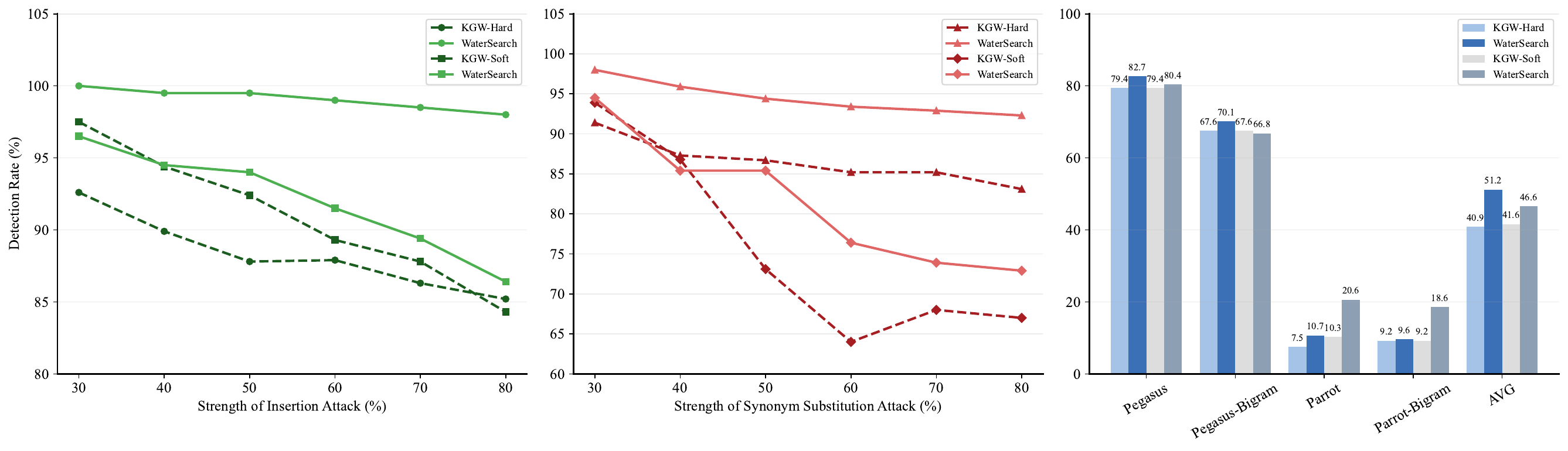}
\caption{Detection success rate under different types of watermark attacks: Insertion attack (left), Synonym substitution attack (middle) and Paraphrase attack (right). WaterSearch exhibits robustness to its counterpart among all the attacks.}
\label{fig:attack}
\end{figure}

Robustness to adversarial perturbations is essential for the practical deployment of watermarking methods. In this section, we evaluate WaterSearch under two categories of attacks: token-level perturbations and paraphrasing.

For token-level attacks, we consider insertion and synonym substitution following prior work~\citep{markllm,simmark}. We generate passages from the RealNewsLike subset of C4~\citep{c4} and apply attacks with strengths ranging from 30\% to 80\%. As shown in Fig.~\ref{fig:attack} (left, middle), WaterSearch consistently maintains higher detection success rates compared to baseline methods across all attack strengths. For paraphrase attacks, which aims to preserve semantic meaning while altering surface forms, we consider two paraphrasers, Pegasus~\citep{pegasus} and Parrot~\citep{parrot}, along with the bigram-based paraphrase attack proposed by \citet{bigram}. As shown in Fig.~\ref{fig:attack} (right), WaterSearch outperforms baseline methods, achieving gains of 10.3 and 9.4 points over KGW-Soft under Parrot and Parrot-Bigram, respectively.

This robustness can be attributed to the aggregation of watermark signals at the sentence and document levels in our detection framework. Even under strong perturbations, sufficient statistical evidence can be accumulated to support reliable hypothesis testing. Full results and additional analysis are provided in Appendix~\ref{appendix:detection}.

\section{Ablation Studies}
\subsection{Parallel Number \texorpdfstring{$\boldsymbol{K}$}{K}}
\label{sec:scaling-k}
\begin{figure}[t]
    \centering

    \begin{subfigure}[t]{0.49\linewidth}
        \centering
        \includegraphics[width=\linewidth]{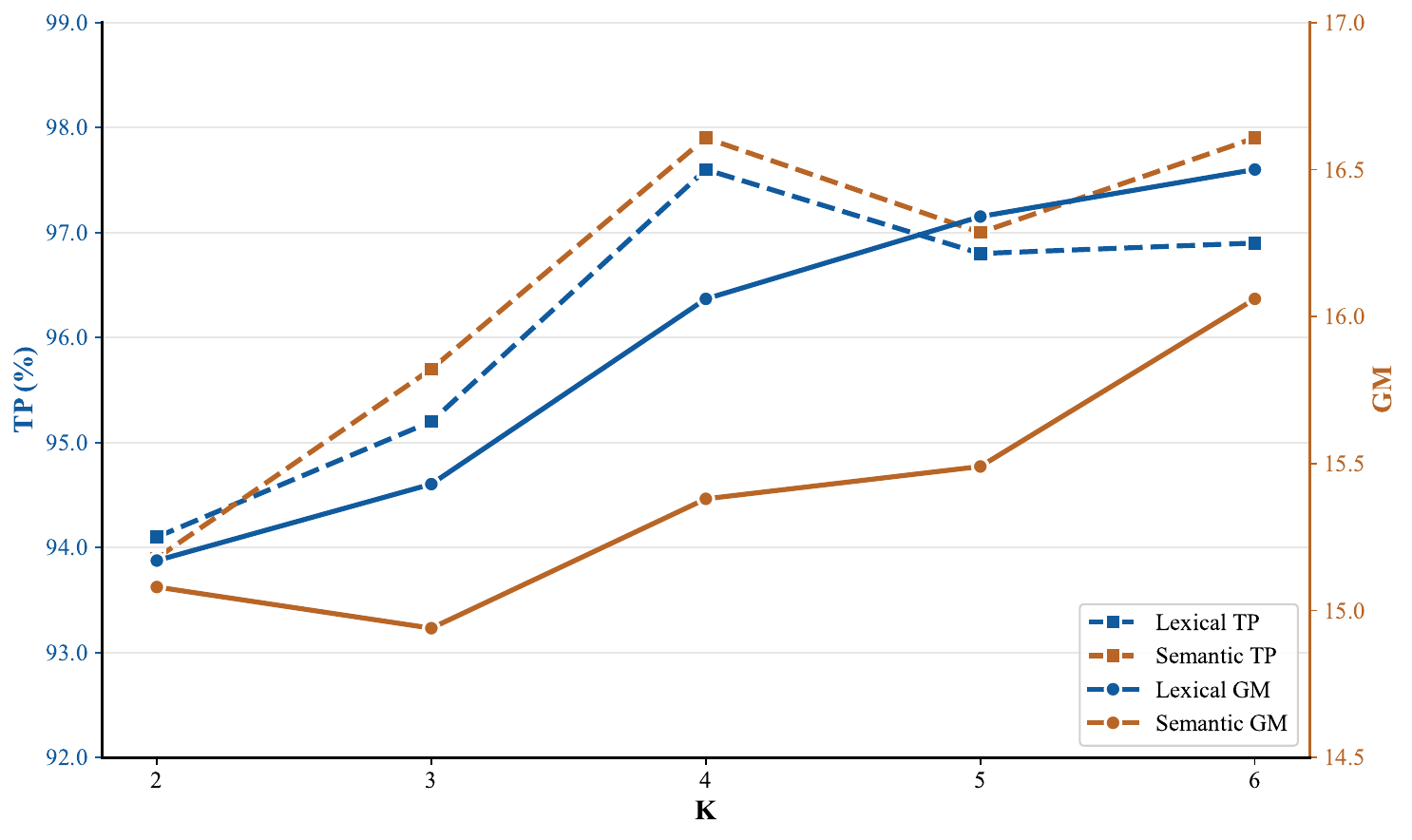}
        \caption{Scaling effect of parallel number $K$.}
        \label{fig:scaling-k}
    \end{subfigure}
    \hfill
    \begin{subfigure}[t]{0.49\linewidth}
        \centering
        \includegraphics[width=\linewidth]{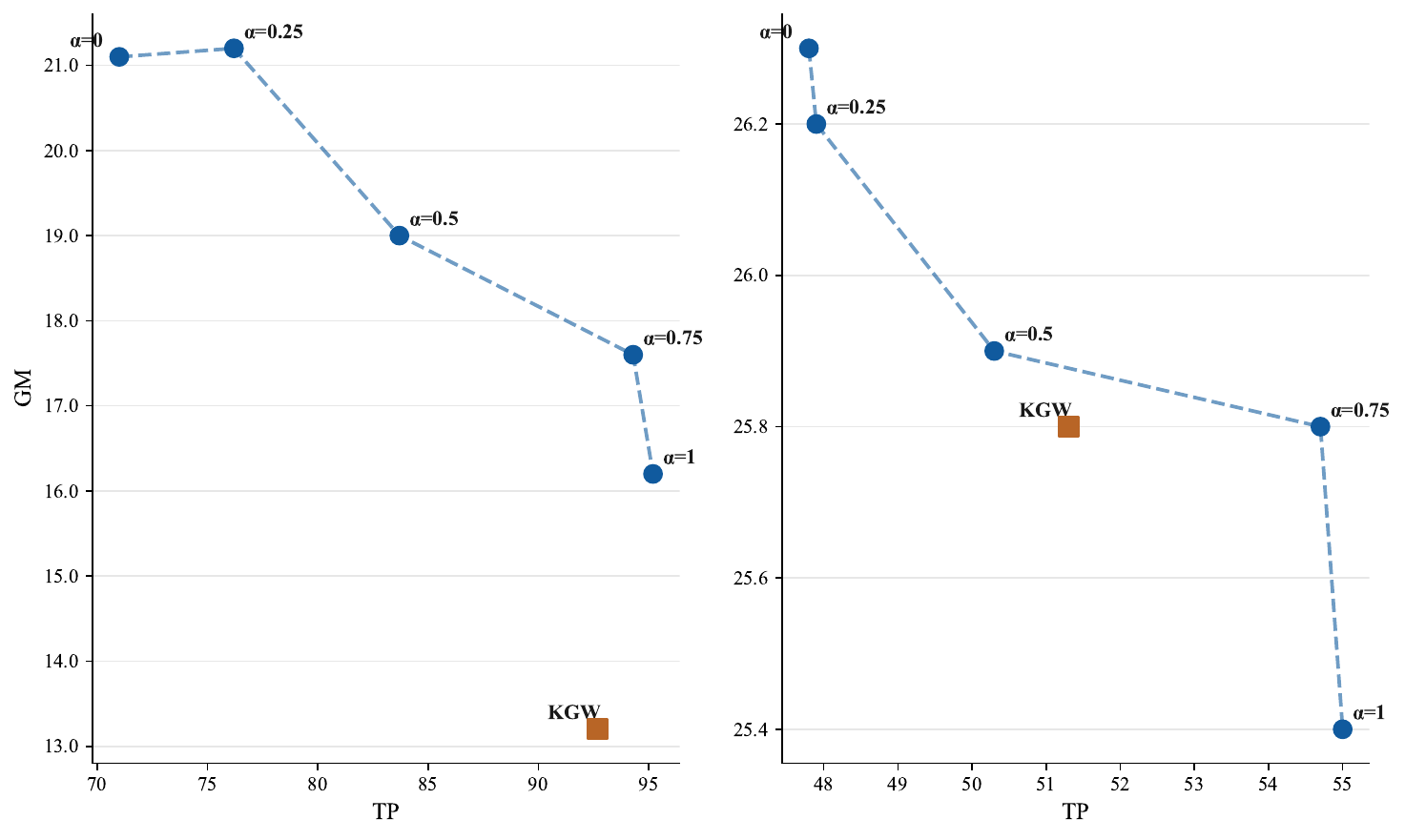}
        \caption{Effect of trade-off factor $\alpha$.}
        \label{fig:tradeoff-alpha}
    \end{subfigure}

    \caption{
    Scaling effect of parallel number $K$ in WaterSearch (left), and the influence of trade-off factor $\alpha$ on text quality and detectability under different watermarking strengths (right).
    }
    \label{fig:scaling-alpha}
\end{figure}

We study the impact of the parallel search size $K$ on generation quality and detection performance. Increasing $K$ expands the candidate pool, leading to better selection outcomes. As shown in Fig.~\ref{fig:scaling-k}, both quality (GM) and detection precision (TP) improve as $K$ increases. Notably, even a small search size ($K=2$) already outperforms the baseline in generation quality, highlighting the effectiveness of seed-level diversity with minimal overhead. Overall, moderate values of $K$ provide a good balance between performance and computational cost. Detailed results are provided in Appendix~\ref{appendix:k}.

\subsection{Chunk Size \texorpdfstring{$\boldsymbol{M}$}{M}}
We examine the effect of chunk size in long-form generation. Results in Table~\ref{tab:ablation_cs} show that larger chunk sizes generally improve performance by reducing semantic drift over long sequences. This suggests that chunk-level selection benefits from longer context when generating extended outputs.



\begin{table}[!htbp]
  \centering
  \small
  \caption{Ablations on different parallel searching methods (Beam number $B=5$).}
  \renewcommand{\arraystretch}{1.2}
  \resizebox{\textwidth}{!}{
    \setlength{\tabcolsep}{4pt}  
    \begin{tabular}{c *{12}{c}}  
\toprule
\textbf{Method}
& \multicolumn{2}{c}{\textbf{Short In/Short Out}} 
& \multicolumn{2}{c}{\textbf{Short In/Long Out}} 
& \multicolumn{3}{c}{\textbf{Long In/Short Out}} 
& \multicolumn{2}{c}{\textbf{Long In/Long Out}} 
& \multicolumn{1}{c}{\textbf{Open-Ended}} 
& \multicolumn{1}{c}{\textbf{Overall}}  \\ 
\cmidrule(lr){2-3} \cmidrule(lr){4-5} \cmidrule(lr){6-8} \cmidrule(lr){9-10} \cmidrule(lr){11-11} \cmidrule(lr){12-12}
& \hspace{6pt}\textbf{KoLA} & \textbf{Copen} & \hspace{8pt}\textbf{ELI5} & \textbf{FiQA} 
& \hspace{2pt}\textbf{HotpotQA} & \textbf{LCC} & \textbf{RepoBench-P}
& \hspace{2pt}\textbf{Multinews} & \textbf{QMsum} 
& \textbf{AlpacaFarm} & \textbf{GM} \\
\midrule
Beam Search
& \hspace{6pt}1.0 & 7.6 & \hspace{8pt}4.4 & 8.7
& 3.0 & 17.4 & 16.7 & 6.9 & 10.1
& 3.4 & 7.9 \\
KGW-Soft (post-hoc)
& \hspace{6pt}\textbf{2.2} & 24.3 & \hspace{8pt}12.4 & 14.7
& \textbf{20.1} & 26.3 & 26.6 & 12.3 & \textbf{13.9}
& \textbf{4.2} & 15.7 \\
\cellcolor{paperblue}WaterSearch
& \cellcolor{paperblue}\hspace{6pt}1.91 & \cellcolor{paperblue}\textbf{27.2} & \cellcolor{paperblue}\hspace{8pt}\textbf{13.2} & \cellcolor{paperblue}\textbf{16.8}
& \cellcolor{paperblue}18.1 & \cellcolor{paperblue}\textbf{27.6} & \cellcolor{paperblue}\textbf{27.1} & \cellcolor{paperblue}\textbf{14.5} & \cellcolor{paperblue}\textbf{13.9}
& \cellcolor{paperblue}3.7 & \cellcolor{paperblue}\textbf{16.4} \\
\hline
\end{tabular}
  }
  \label{tab:ablation-parallel-method}
\end{table}
\begin{table}[htbp]
\centering
\small
\caption{Ablation on chunk size (tokens) of long output tasks.}
\begin{tabular}{cccccc}
\toprule
 Task & Metric & 20 & 40 & 60 & 80 \\
\midrule
\multirow{2}{*}{MultiNews} & GM & 13.62 & 13.84 & 13.71 & 13.92 \\
 & TP & 100 & 100 & 100 & 100 \\
\multirow{2}{*}{QMSum} & GM & 13.49 & 13.59 & 13.63 & 13.54 \\
 & TP & 100 & 100 & 100 & 100 \\
\bottomrule
\end{tabular}
\label{tab:ablation_cs}
\end{table}

\subsection{Balance Text Quality and Detectability by Factor \texorpdfstring{$\boldsymbol{\alpha}$}{alpha}}
We analyze the effect of the trade-off parameter $\alpha$ on text quality and watermark detectability. Experiments are conducted on four datasets (Copen, ELI5, HotpotQA, and MultiNews) using Llama-2-7B-Chat under both low and high watermarking strength. Fig.~\ref{fig:tradeoff-alpha} show that increasing $\alpha$ strengthens watermark detectability while gradually reducing text quality across different strength. Despite this trade-off, WaterSearch consistently achieves a better Pareto frontier than baseline methods, with more pronounced gains under stronger watermarking regimes. Full results can be found from Fig.~\ref{fig:tradeoff-high-rougel} to Fig.~\ref{fig:tradeoff-low-semsim} in Appendix~\ref{appendix:alpha}.

\subsection{Comparison with Parallel Generation Strategies}
We compare WaterSearch with alternative parallel generation strategies, including beam search and rejection sampling, under watermarking settings. Beam search, which optimizes token-level likelihood, tends to propagate early distortions introduced by watermark perturbations. Rejection sampling improves diversity through repeated sampling, but remains limited to a fixed watermarking scheme and cannot capture variations induced by different watermark priors. In contrast, WaterSearch introduces diversity at the seed level, where each seed corresponds to a distinct watermark perturbation. Moreover, chunk-wise selection with dynamic context updating mitigates the accumulation of watermark-induced distortions over long sequences. As shown in Table~\ref{tab:ablation-parallel-method}, WaterSearch consistently outperforms both approaches, with particularly strong gains in long-form generation, highlighting the effectiveness of seed-level exploration.

\vspace{-10pt}

\section{Computation Cost}
WaterSearch introduces additional computational overhead due to parallel candidate generation. A naive implementation with $K$ parallel branches would require $\mathcal{O}(K)$ memory and reduce throughput proportionally. However, the actual cost is significantly mitigated by two key design choices. First, modern LLM architectures support efficient parallel inference, allowing multiple decoding branches to be processed in a single forward pass. Second, WaterSearch leverages KV-cache reuse across chunks, reducing the peak memory complexity from $\mathcal{O}(k(L+T))$ to $\mathcal{O}(k(L+m))$, where $m \ll T$ is the chunk size.

\begin{figure}[t!]
    \centering
    \includegraphics[width=0.8\linewidth]{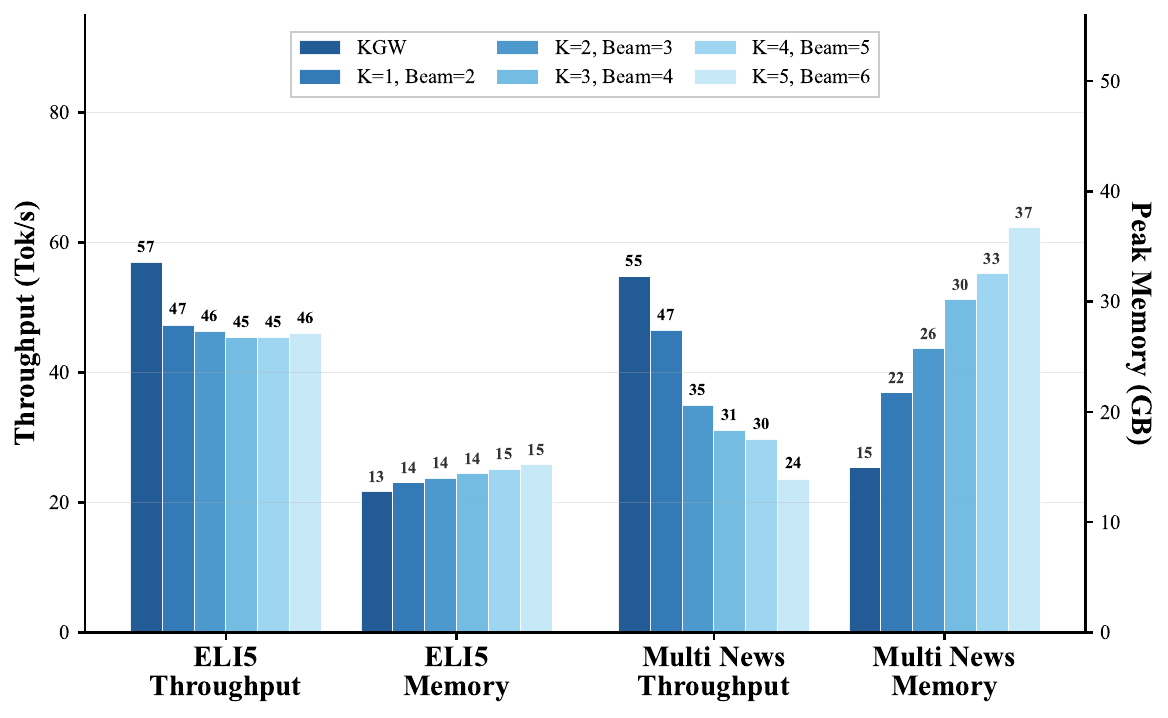}
    \caption{Throughput and peak memory usage of WaterSearch.}
    \label{fig:throughput}
\end{figure}
\vspace{-10pt}
We empirically evaluate throughput and memory usage on representative short- and long-context generation tasks. As shown in Fig.~\ref{fig:throughput}, WaterSearch incurs moderate overhead compared to standard decoding. For short-input scenarios, the impact on throughput and memory is minimal, while for long-input tasks, the overhead becomes more noticeable but remains sub-linear with respect to $K$, with strong performance gains already achievable at small $k$ (e.g., $k=2$).

\section{Conclusion}
In this work, we revisit LLM watermarking from a seed-level perspective and propose WaterSearch, a search-based framework that leverages seed pooling and candidate selection to improve the quality--detectability trade-off. We further develop a detection method based on seed-level statistical variation, enabling robust hypothesis testing without explicit seed identification. Experiments show consistent improvements across tasks, especially in challenging scenarios such as low-entropy and short-text generation, highlighting the effectiveness of treating randomness as a controllable factor in watermarking.

\section*{LIMITATIONS}
This paper proposes WaterSearch, a watermarking framework that improves text quality through parallel search and enhances diversity via seed pooling. As a modular framework, WaterSearch inherits both the strengths and limitations of the underlying watermarking methods.

For instance, KGW-based methods introduce watermark signals by modifying token-level probability distributions, which may still affect generation quality. Semantic-based methods often rely on rejection sampling, leading to higher computational cost. In addition, although our parallel generation process is optimized, WaterSearch still incurs additional computational overhead compared to standard decoding. Improving sampling efficiency and reducing computational cost remain important directions for future work.

\section*{ETHICS STATEMENT}

We conducted all experiments using open-source datasets and did not involve any personally identifiable information or sensitive content. The primary goal of our research is to alleviate the misuse of AI models—a pursuit with significant potential benefits for scientific research and engineering.

\section*{REPRODUCIBILITY STATEMENT}

To support reproducibility, we provide detailed descriptions of our metrics and framework in Section \ref{sec:method} and Appendix \ref{pseudo-code}. Additionally, we include comprehensive implementation details and hyperparameter selection guidelines in Section \ref{ssec:exp-setup} and Appendix \ref{appendix:params}. We will release our data and code following the anonymous review process. Our implementation is built upon the WaterBench codebase~\citep{waterbench}. For adversarial robustness evaluation, we adopt the token-level attack implementation from MarkLLM~\citep{markllm} and the paraphrase attack codebase from SimMark~\citep{simmark}.

\bibliography{iclr2026_conference}
\bibliographystyle{iclr2026_conference}

\clearpage
\appendix

\section*{APPENDIX}





\section{Seed Generation Process and Pseudo-Code of WaterSearch}
\label{pseudo-code}

The seed generation process in WaterSearch consists of two components: (1) seed pool generation and (2) token-level seed generation.

\paragraph{Seed Pool Construction.}
To enable diverse watermark behaviors while ensuring reproducibility for detection, we introduce a seed pool mechanism that maintains a set of candidate seeds at each generation step. Specifically, we construct the seed pool via a key-conditioned permutation:
\begin{equation}
\mathcal{S} = \pi_{\text{key}}([s_1, s_2, \cdots]),
\label{eq:seed_pool}
\end{equation}
where $[s_1, s_2, \cdots]$ is a predefined seed set, and $\pi_{\text{key}}(\cdot)$ denotes a permutation function parameterized by a secret key. This ensures that the same seed ordering can be reconstructed during detection given the key.

At each generation step, we select a subset of seeds $\{s_i\}_{i=1}^{K} \subset \mathcal{S}$ to instantiate parallel candidates. For each candidate, the token-level seed is derived from the context following Eq.~(\ref{eq:hash_func}), with $s_i$ serving as the global seed controlling the watermark partition. 

In practice, $\pi_{\text{key}}(\cdot)$ can be implemented as a key-dependent shuffle. This design introduces controlled variation in watermark behavior across candidates while remaining compatible with existing token-level watermarking methods.
\paragraph{Token-level Seed Generation.}
Following the KGW framework, the seed at each decoding step is derived from recent context:
\begin{equation}
\begin{aligned}
s_t 
&= \text{Hash}(x_{t-h}, \cdots, x_{t-1}),
\end{aligned}
\label{eq:seed}
\end{equation}
which can be implemented in practice as:
\begin{equation}
s_t = \Big(\prod_{i=1}^{h} x_{t-i}\Big) \bmod M,
\end{equation}
where $M$ is a large prime number~\citep{kgw,three-bricks}.

Based on Eq.~\ref{eq:seed_pool} and Eq.~\ref{eq:seed}, we present the pseudo-code for generation and detection in Algorithm~\ref{alg:pseudo-generation} and Algorithm~\ref{alg:pseudo-detection}.
\begin{algorithm}[htbp]
\small
\caption{Pseudo-Code of WaterSearch Generation}
\label{alg:pseudo-generation}
\begin{algorithmic}[1]

\State \textbf{Input:} prompt $\mathbf{c}$, chunk size $m$, parallel size $K$
\State \textbf{Init:} random.seed(key), $seed\_pool \gets [1,\dots,N]$

\For{$i = 1$ to $\lceil T/m \rceil$}
    \State \textbf{// Seed Pool Shuffle}
    \State random.shuffle($seed\_pool$)
    \State $cur\_seeds \gets seed\_pool[1:K]$
    \Comment{Eq.~(\ref{eq:seed_pool})}
    \State \textbf{// Reference Generation}
    \State $y_i^{(0)} \gets$ model.generate($\mathbf{c}$)
    \Comment{unwatermarked chunk}
    \State \textbf{// Watermarked Candidates}
    \State $wm\_chunks \gets [\,]$
    \For{$j = 1$ to $K$}
        \State $s_t \gets$ hash\_func($\mathbf{c}[-h:],~cur\_seeds[j]$)
        \Statex\Comment{Eq.~(\ref{eq:hash_func})}
        \State $y_i^{(j)} \gets$ model.generate($\mathbf{c}$, $wm\_processor$, $s_t$)
        \State Append($wm\_chunks$, $y_i^{(j)}$)
    \EndFor
    \State \textbf{// Candidate Selection}
    \State $j^* \gets$ calculate\_best($y_i^{(0)}, wm\_chunks$)
    \Comment{Eq.~(\ref{eq:weighted})}
    \State \textbf{// Context Update}
    \State $\mathbf{c} \gets [\mathbf{c} \,\|\, wm\_chunks[j^*]]$
\EndFor
\State \textbf{Output:} tokenizer.decode($\mathbf{c}$)
\end{algorithmic}
\end{algorithm}

\begin{algorithm}[htbp]
\small
\caption{Pseudo-Code of WaterSearch Detection}
\label{alg:pseudo-detection}
\begin{algorithmic}[1]

\State \textbf{Input:} response $\mathbf{y}$, chunk size $m$, parallel size $K$
\State \textbf{Init:} random.seed(key), $seed\_pool \gets [1,\dots,N]$
\State $\mathcal{P} \gets [\,]$

\State $\{\mathbf{y}_1,\dots,\mathbf{y}_{\lceil T/m\rceil}\} \gets \text{Split}(\mathbf{y}, m)$

\For{$i = 1$ to $\lceil T/m\rceil$}

    \State \textbf{// Seed Reconstruction}
    \State random.shuffle($seed\_pool$)
    \State $cur\_seeds \gets seed\_pool[1:K]$
    \Comment{Eq.~(\ref{eq:seed_pool})}

    \State \textbf{// Chunk Test}
    \State $p_i \gets$ chunk\_detect($\mathbf{y}_i$, $cur\_seeds$)
    \Comment{Eq.~(\ref{eq:p-val})}

    \State $\mathcal{P}.\text{append}(p_i)$

\EndFor

\State \textbf{// Aggregation}
\State $p_{\text{doc}} \gets$ document\_detect($\mathcal{P}$)
\Comment{Eq.~(\ref{eq:fisher_combined})}

\State \textbf{Output:} $p_{\text{doc}}$

\end{algorithmic}
\end{algorithm}

\section{Connections between Token-level and Chunk-level Similarity}
\label{appendix:theory}

In this section, we analyze the relationship between token-level watermark optimization and chunk-level selection used in WaterSearch. Our goal is to show that, under reasonable assumptions, the macroscopic selection objective aligns with the microscopic watermarking objective.

\paragraph{Macroscopic Objective.}
In WaterSearch, candidate selection is performed at the chunk level by maximizing a weighted objective:
\begin{equation}
    \mathbf{q}(y,\tilde{y})=\alpha\mathbf{q}_\text{sen-sim}(y,\tilde{y}) + (1-\alpha)\frac{|\tilde{y}|_G}{|\tilde{y}|},
\end{equation}
which balances semantic similarity and watermark detectability.

\paragraph{Microscopic Objective.}
At the token level, prior work~\citep{morphmark} formulates watermarking as a trade-off between distributional fidelity and detectability:
\begin{equation}
\mathcal{J}(r) = \mathcal{T}(r) + \omega \cdot \mathcal{W}(r),
\end{equation}
where $r$ denotes watermark strength, $\mathcal{T}(r)$ measures deviation from the original distribution, and $\mathcal{W}(r)$ captures watermark effectiveness.

\paragraph{Proposition.}
\textit{
Under the assumptions below, the macroscopic objective $\mathbb{E}[\mathbf{q}(y,\tilde{y})]$ and the microscopic objective $\mathcal{J}(r)$ share the same maximizer $r^*$ when the weight is chosen as
\begin{equation}
\omega=\frac{1-\alpha}{2\alpha}\cdot\frac{1}{f'(\mathcal{T}(r^*))},
\end{equation}
where $f(\cdot)$ maps distribution-level similarity to semantic similarity.
}

\paragraph{Assumptions.}
For analytical tractability, we make the following assumptions:

\begin{itemize}
    \item \textbf{Token independence:} tokens are sampled independently from the perturbed distribution $\hat{P}$.
    \item \textbf{Detectability consistency:} the expected fraction of green tokens satisfies $\mathbb{E}\left[\frac{|\tilde{y}|_G}{|\tilde{y}|}\right] = \hat{P}_G$.
    \item \textbf{Semantic proxy:} expected semantic similarity is a smooth increasing function of distribution similarity, i.e., $\mathbb{E}[\mathbf{q}_\text{sen-sim}] = f(\mathcal{T}(r))$.
\end{itemize}

\paragraph{Sketch of Proof.}
Under the above assumptions, the macroscopic objective becomes:
\begin{equation}
\mathbb{E}[\mathbf{q}] = \alpha f(\mathcal{T}(r)) + (1-\alpha)\hat{P}_G.
\end{equation}

Taking derivatives with respect to $r$ and comparing with the stationary condition of $\mathcal{J}(r)$ yields a consistent optimal solution when $\omega$ is set accordingly. The key observation is that both objectives balance the same two quantities: distributional fidelity and watermark signal strength.

\paragraph{Instantiation under KGW.}
For KGW-style watermarking, the green token probability satisfies:
\begin{equation}
\hat{P}_G = P_G + r(1-P_G), \quad 
\mathcal{W}(r)=2r(1-P_G),
\end{equation}
which leads to:
\begin{equation}
\frac{d\hat{P}_G}{dr} = 1-P_G, \quad 
\frac{d\mathcal{W}}{dr} = 2(1-P_G).
\end{equation}

Substituting into the optimality condition gives:
\begin{equation}
\omega=\frac{1-\alpha}{2\alpha}\cdot\frac{1}{f'(\mathcal{T}(r^*))}.
\end{equation}

\paragraph{Discussion.}
This result suggests that chunk-level selection in WaterSearch provides a principled approximation to token-level watermark optimization. Although the assumptions are simplified, the analysis offers theoretical support for using sentence-level search to balance quality and detectability in practice.

\section{Dataset Details}
The information of all the datasets is shown in Table \ref{tab:data_intro}, which categorizes based on the length of input or output. Apart from WaterBench~\citep{waterbench}, we additionally introduce RepoBench-P~\citep{repobench} to complete evaluations in low-entropy scenarios.

\newcolumntype{C}{>{\centering\arraybackslash}X}

\begin{table*}[!htbp]
  \centering
  \small
  \renewcommand{\arraystretch}{1.2}
  \caption{Details of all the experiment datasets.}
\resizebox{\textwidth}{!}{
  \begin{tabular}{l l l r c c}
    \toprule
    \textbf{Category} & \textbf{Dataset} & \textbf{Task} & \textbf{Metric} & \textbf{Avg.Length} \\
    \midrule
    \addlinespace[4pt]
    \multirow{2}{*}[0em]{\makecell[c]{\textit{Short In/Short Out}}}
    & KoLA~\citep{kola} & Entity Probing & F1 & 11.9/5.9 & \\
    \addlinespace[2pt]
    & Copen~\citep{copen} & Concept Probing & F1 & 84.1/3.9 & \\
    \addlinespace[4pt]
    
    \multirow{2}{*}[0em]{\makecell[c]{\textit{Short In/Long Out}}}
    & ELI5~\citep{eli5} & Long-form QA & ROUGE-L & 49.4/277.2 & \\
    \addlinespace[2pt]
    & FiQA~\citep{fiqa} & Finance QA & ROUGE-L & 17.9/302.2 & \\
    \addlinespace[4pt]
    
    \multirow{3}{*}[0em]{\makecell[c]{\textit{Long In/Short Out}}}
    & HotpotQA~\citep{hotpotqa} & Multi-Doc QA & F1 & 15266.0/5.6 & \\
    \addlinespace[2pt]
    & LCC~\citep{lcc} & Code Completion & Edit Distance & 4183.8/16.8 & \\
    \addlinespace[4pt]
    & RepoBench-P~\citep{repobench-p} & Code Completion & Edit Distance & 14696.8/18.9 & \\
    \addlinespace[4pt]
    
    \multirow{2}{*}[0em]{\makecell[c]{\textit{Long In/Long Out}}}
    & Multinews~\citep{multinews} & Multi-Doc Summary & ROUGE-L & 3114.4/329.7 & \\
    \addlinespace[2pt]
    & QMsum~\citep{qmsum} & Query-Based Summary & ROUGE-L & 15923.1/87.9 & \\
    \addlinespace[4pt]

    \multirow{2}{*}[0em]{\makecell[c]{\textit{Open-Ended Generation}}}
    & AlpacaFarm~\citep{alpacafarm} & Instruction Following & GPT-4 Judge & 41.0/86.2 & \\
    \addlinespace[2pt]
    & C4\citep{c4} & General Writing & Perplexity & 35.4/233.2 & \\
    
    \bottomrule
  \end{tabular}
  }
  \label{tab:data_intro}
\end{table*}



\section{More Experimental Results}

\subsection{Detailed Experimental Setups}\label{appendix:params}
For the of baseline method in Table \ref{tab:main-result}, we report the result in \citep{waterbench}. Considering the timeliness of the models, we have added Qwen-2.5-Instruct as a supplement. Aligned with WaterBench settings, the $\gamma,\  \delta$ hyper-paramters are acquired by grid search to reach 0.95 True Positive Rate. We first initialize the hyper-parameters for the Qwen-2.5 model experiments using those from the Llama-2 model experiments. As an increase in $\gamma$ leads to weaker watermark strength and an increase in $\delta$ results in stronger watermarking, we leveraged this relationship during grid search to find the hyper-parameter setting that minimizes the smallest deviation from the target TPR. The full settings are shown in \ref{appendix:tab:params}.

\begin{table}[!htbp]
  \centering
  \small
    \setlength{\tabcolsep}{3pt}
    \begin{tabular}{lccc}
      \toprule
      \textbf{Method / Params} & \textbf{Llama2-7B-Chat} & \textbf{InternLM-7B-Chat} & \textbf{Qwen2.5-7B-Instruct} \\
      \midrule
      Hard Watermark ($\gamma$) & 0.25 & 0.15 & 0.35 \\
      Soft Watermark ($\gamma$/$\delta$) & 0.1/10 & 0.1/10 & 0.1/12 \\
      GPT Watermark ($\gamma$/$\delta$) & 0.1/10 & 0.25/15 & 0.1/12 \\
      V2 Watermark ($\gamma$/$\delta$) & 0.25/15 & 0.1/10 & 0.25/15 \\
      \bottomrule
    \end{tabular}
  \caption{Hyper-parameters settings of baseline methods}
  \label{appendix:tab:params}
\end{table}

As for WaterSearch method, since the watermarking strength is stronger than KGW-based methods, we fixed $\gamma$ the same as counterpart, and decrease $\delta$ to reach the overall 95\% TPR. After that, we turn $\alpha$ for a better performance, which is further discussed in Appendix \ref{appendix:alpha}. 

\subsection{Experimental Results on More Models}
We evaluate WaterSearch on more models, including state-of-the-art models Qwen3-14B and Qwen3-32B~\citep{qwen3}. Here we set $\gamma=0.1,\delta=12.0$ in KGW-Soft and  $\gamma=0.05,\delta=12.0$ in WaterSearch to assure the watermarking strength around 95\%. The results are shown in Table \ref{tab:qwen3-14b} and Table \ref{tab:qwen3-32b}, which demonstrate the effectiveness across diverse models. During experiment, we notice the strong LLMs may possess a sharp logit distribution, which worths for further discovering. Limited by the maximum memory of single graphics card, we only report the result of Qwen3-14B in 4K and Qwen3-32B in 3K context length. We will support multi-device implementation to facilate LLMs in larger scale.

\begin{table}[htbp]
  \centering
  \caption{Comparison of WaterSearch and KGW-soft on Qwen-3-14B in 4K context length.}
  \small
  \renewcommand{\arraystretch}{1.2}
  \resizebox{\textwidth}{!}{
    \setlength{\tabcolsep}{4pt}  
    \begin{tabular}{c *{13}{c}}  
\toprule
\textbf{Method}
& \multicolumn{2}{c}{\textbf{Short In/Short Out}} 
& \multicolumn{2}{c}{\textbf{Short In/Long Out}} 
& \multicolumn{3}{c}{\textbf{Long In/Short Out}} 
& \multicolumn{2}{c}{\textbf{Long In/Long Out}} 
& \multicolumn{1}{c}{\textbf{Open-Ended}} 
& \multicolumn{2}{c}{\textbf{Overall}}  \\ 
\cmidrule(lr){2-3} \cmidrule(lr){4-5} \cmidrule(lr){6-8} \cmidrule(lr){9-10} \cmidrule(lr){11-11} \cmidrule(lr){12-13}
& \hspace{6pt}\textbf{KoLA} & \textbf{Copen} & \hspace{8pt}\textbf{ELI5} & \textbf{FiQA} 
& \hspace{2pt}\textbf{HotpotQA} & \textbf{LCC} & \textbf{RepoBench-P}
& \hspace{2pt}\textbf{Multinews} & \textbf{QMsum} 
& \textbf{AlpacaFarm} & \textbf{TP} & \textbf{GM} \\
\midrule
KGW-Soft
& \hspace{6pt}12.6 & 51.6 & \hspace{8pt}16.5 & 15.3
& 43.4 & 24.6 & 24.4 & 13.4 & 16.9
& 16.9 & 93.1 & 23.6 \\
\cellcolor{paperblue}WaterSearch
& \cellcolor{paperblue}\hspace{6pt}13.5 & \cellcolor{paperblue}51.6 & \cellcolor{paperblue}\hspace{8pt}20.9 & \cellcolor{paperblue}17.4
& \cellcolor{paperblue}50.5 & \cellcolor{paperblue}36.0 & \cellcolor{paperblue}28.6 & \cellcolor{paperblue}50.5 & \cellcolor{paperblue}20.1
& \cellcolor{paperblue}25.3 & \cellcolor{paperblue}91.4  & \cellcolor{paperblue}\textbf{31.4} \\
\bottomrule
\end{tabular}
  }
  \label{tab:qwen3-14b}
\end{table}
\begin{table}[htbp]
  \centering
  \small
  \caption{Comparison of WaterSearch and KGW-soft on Qwen-3-32B in 3K context length.}
  \renewcommand{\arraystretch}{1.2}
  \resizebox{\textwidth}{!}{
    \setlength{\tabcolsep}{4pt}  
    \begin{tabular}{c *{13}{c}}  
\toprule
\textbf{Method}
& \multicolumn{2}{c}{\textbf{Short In/Short Out}} 
& \multicolumn{2}{c}{\textbf{Short In/Long Out}} 
& \multicolumn{3}{c}{\textbf{Long In/Short Out}} 
& \multicolumn{2}{c}{\textbf{Long In/Long Out}} 
& \multicolumn{1}{c}{\textbf{Open-Ended}} 
& \multicolumn{2}{c}{\textbf{Overall}}  \\ 
\cmidrule(lr){2-3} \cmidrule(lr){4-5} \cmidrule(lr){6-8} \cmidrule(lr){9-10} \cmidrule(lr){11-11} \cmidrule(lr){12-13}
& \hspace{6pt}\textbf{KoLA} & \textbf{Copen} & \hspace{8pt}\textbf{ELI5} & \textbf{FiQA} 
& \hspace{2pt}\textbf{HotpotQA} & \textbf{LCC} & \textbf{RepoBench-P}
& \hspace{2pt}\textbf{Multinews} & \textbf{QMsum} 
& \textbf{AlpacaFarm} & \textbf{TP} & \textbf{GM} \\
\midrule
KGW-Soft
& \hspace{6pt}9.3 & 43.4 & \hspace{8pt}11.4 & 11.1
& 34.5 & 14.0 & 14.0 & 9.7 & 15.5
& 4.0 & 97.4 & 16.7 \\
\cellcolor{paperblue}WaterSearch
& \cellcolor{paperblue}\hspace{6pt}9.4 & \cellcolor{paperblue}42.7 & \cellcolor{paperblue}\hspace{8pt}19.9 & \cellcolor{paperblue}16.9
& \cellcolor{paperblue}41.0 & \cellcolor{paperblue}14.4 & \cellcolor{paperblue}13.9 & \cellcolor{paperblue}15.0 & \cellcolor{paperblue}16.1
& \cellcolor{paperblue}7.3 & \cellcolor{paperblue}95.9 & \cellcolor{paperblue}\textbf{19.7} \\
\bottomrule
\end{tabular}
  }
  \label{tab:qwen3-32b}
\end{table}







\subsection{Comparison on Lexical-based and Semantic-based Methods}\label{appendix:metric}
\begin{table*}[!htbp]
  \centering
  \small
  \renewcommand{\arraystretch}{1.2}
  \caption{Lexical-sequential and semantic similarity for chunk selection.}
    \resizebox{\textwidth}{!}{
    \begin{tabular}{lccccccccccc}
    \toprule
    Similarity Metric & KoLA & Copen & ELI5 & FiQA & HotpotQA & LCC & RepoBench-P & Multinews & QMsum & AlpacaFarm & Avg  \\
    \midrule
    Lexical  & \textbf{2.4}  & \textbf{27.2}  & 12.4 & 16.0   & 16.3   & \textbf{27.6} & \textbf{27.1}  & 13.6  & 13.5  & \textbf{4.5}        & \textbf{16.1} \\
    Semantic &  1.9    & \textbf{27.2}  & \textbf{13.2}  &  \textbf{16.8}   & \textbf{18.1}  & 22.8 & 20.7 &  \textbf{14.5} & \textbf{13.9} & 3.7 & 15.3 \\
    \bottomrule
    \end{tabular}
}
\label{tab:lexical-semantic}
\end{table*}
This section compares the impact of lexical and semantic similarity metrics, using Rouge-L~\citep{rouge-l} and Sentence-BERT~\citep{s-bert}, respectively. As shown in Table~\ref{tab:lexical-semantic}, Rouge-L achieves higher average performance overall, but the relative effectiveness varies across task types. For short-answer tasks such as KoLA and Copen, the two metrics yield comparable results. In contrast, for comprehension-oriented tasks such as HotpotQA, MultiNews, and QMSum, Sentence-BERT performs better, as it captures semantic information in a vector space and evaluates similarity via cosine distance. For code completion tasks such as LCC and RepoBench-P, Sentence-BERT underperforms due to the lack of domain-specific training on source code, whereas longest common subsequence matching is more naturally aligned with the structural properties of code.

\section{Detection of WaterSearch}
\label{appendix:detection}

While our sentence-level partitioning may seem vulnerable to targeted attacks (e.g., modifying sentence-final tokens), we emphasize that this vulnerability assumes the partitioning scheme is known to attackers. In practice, the partitioning strategy can be designed to remain agnostic to external observers, effectively mitigating such targeted attacks.

\begin{table}[htbp]
  \centering
  \small
  \caption{Comparison on Insert Attack and Synonym Substitution Attack.}
  \begin{tabular}{clcccccc}
    \toprule
    Attack Type & Method & 0.3 & 0.4 & 0.5 & 0.6 & 0.7 & 0.8 \\
    \midrule
    \multirow{4}{*}{Insert Attack} 
    & KGW-Hard & 92.6 & 89.9 & 87.8 & 87.9 & 86.3 & 85.2 \\
    & \cellcolor{paperblue}WaterSearch & \cellcolor{paperblue}100.0 & \cellcolor{paperblue}99.5 & \cellcolor{paperblue}99.5 & \cellcolor{paperblue}99.0 & \cellcolor{paperblue}98.5 & \cellcolor{paperblue}98.0 \\
    & KGW-Soft & 97.5 & 94.4 & 92.4 & 89.3 & 87.8 & 84.3 \\
    & \cellcolor{paperblue}WaterSearch & \cellcolor{paperblue}96.5 & \cellcolor{paperblue}94.5 & \cellcolor{paperblue}94.0 & \cellcolor{paperblue}91.5 & \cellcolor{paperblue}89.4 & \cellcolor{paperblue}86.4 \\
    \multirow{4}{*}{SynSub Attack} 
    & KGW-Hard & 91.4 & 87.3 & 86.7 & 85.2 & 85.2 & 83.1 \\
    & \cellcolor{paperblue}WaterSearch & \cellcolor{paperblue}98.0 & \cellcolor{paperblue}95.9 & \cellcolor{paperblue}94.4 & \cellcolor{paperblue}93.4 & \cellcolor{paperblue}92.9 & \cellcolor{paperblue}92.3 \\
    & KGW-Soft & 93.9 & 86.8 & 73.1 & 64.0 & 68.0 & 67.0 \\
    & \cellcolor{paperblue}WaterSearch & \cellcolor{paperblue}94.5 & \cellcolor{paperblue}85.4 & \cellcolor{paperblue}85.4 & \cellcolor{paperblue}76.4 & \cellcolor{paperblue}73.9 & \cellcolor{paperblue}72.9 \\
    \bottomrule
  \end{tabular}
  \label{tab:token-attack}
\end{table}

\begin{table}[htbp]
  \centering
  \small
  \caption{Comparison on Paraphrase-Based Attacks.}
  \begin{tabular}{lccccc}
    \toprule
    Method & Pegasus & Pegasus-Bigram & Parrot & Parrot-Bigram & AVG \\
    \midrule
    KGW-Hard 
      & 79.4 & 67.6 & 7.5 & 9.2 & 40.9 \\

    \cellcolor{paperblue}WaterSearch 
      & \cellcolor{paperblue}90.2
      & \cellcolor{paperblue}74.2
      & \cellcolor{paperblue}20.6
      & \cellcolor{paperblue}19.6
      & \cellcolor{paperblue}\textbf{51.2} \\

    KGW-Soft 
      & 79.4 & 67.6 & 10.3 & 9.2 & 41.6 \\

    \cellcolor{paperblue}WaterSearch 
      & \cellcolor{paperblue}80.4
      & \cellcolor{paperblue}66.8
      & \cellcolor{paperblue}20.6
      & \cellcolor{paperblue}18.6
      & \cellcolor{paperblue}\textbf{46.6} \\

    \bottomrule
  \end{tabular}
  \label{tab:para-attack}
\end{table}

\section{Ablation Study}\label{ablation}

\subsection{Ablation on Parallel Number $K$}\label{appendix:k}
This part show the full experimental results of ablation on parallel number $K$. The experiment is conducted on Llama2-7B-Chat with $\gamma=0.15,\delta=10,\alpha=0.75$. Text quality is judged by lexical and semantic similarity, respectively. Table \ref{tab:scaling-k-lexical} and Table \ref{tab:scaling-k-semantic} demonstrate the existance of Test time Scaling Law in both task completeness and detectability. 
\begin{table}[htbp]
  \centering
  \small
  \renewcommand{\arraystretch}{1.2}
  \caption{Scaling effect of WaterSearch in Generation and Detection by lexical similarity metric.}
  \resizebox{\textwidth}{!}{
    \setlength{\tabcolsep}{4pt}  
    \begin{tabular}{c *{13}{c}}  
\toprule
\textbf{Config}
& \multicolumn{2}{c}{\textbf{Short In/Short Out}} 
& \multicolumn{2}{c}{\textbf{Short In/Long Out}} 
& \multicolumn{3}{c}{\textbf{Long In/Short Out}} 
& \multicolumn{2}{c}{\textbf{Long In/Long Out}} 
& \multicolumn{1}{c}{\textbf{Open-Ended}} 
& \multicolumn{2}{c}{\textbf{Overall}}  \\ 
\cmidrule(lr){2-3} \cmidrule(lr){4-5} \cmidrule(lr){6-8} \cmidrule(lr){9-10} \cmidrule(lr){11-11} \cmidrule(lr){12-13}
& \hspace{6pt}\textbf{KoLA} & \textbf{Copen} & \hspace{8pt}\textbf{ELI5} & \textbf{FiQA} 
& \hspace{2pt}\textbf{HotpotQA} & \textbf{LCC} & \textbf{RepoBench-P}
& \hspace{2pt}\textbf{Multinews} & \textbf{QMsum} 
& \textbf{AlpacaFarm} & \textbf{TP} & \textbf{GM} \\
\midrule
K=2
& 1.9 & 27.1 & \hspace{8pt}13.5 & 15.7
& 17.0 & 24.1 & 22.6 & 13.5 & 13.5
& 2.8 & 94.1 & 15.2 \\
K=3 
& 2.4 & 24.0 & \hspace{8pt}12.5 & 15.9
& 18.6 & 26.3 & 23.3 & 13.5 & 13.2
& 4.6 & 95.2 & 15.4 \\
K=4 
& 2.4 & 27.2 & \hspace{8pt}12.4 & 16.0
& 16.3 & 27.6 & 27.1 & 13.6 & 13.5
& 4.5 & 97.6 & 16.1 \\
K=5 
& 3.1 & 26.5 & \hspace{8pt}15.6 & 16.1
& 15.5 & 28.2 & 26.5 & 13.9 & 13.6
& 4.4 & 96.8 & 16.3 \\
K=6 
& 3.0 & 27.3 & \hspace{8pt}12.5 & 16.4
& 17.8 & 29.4 & 27.1 & 13.7 & 13.5
& 4.3 & 96.9 & 16.5 \\
\bottomrule
\end{tabular}
  }
  \label{tab:scaling-k-lexical}
\end{table}
\vspace{-3pt}
\begin{table}[htbp]
  \centering
  \small
  \renewcommand{\arraystretch}{1.2}
  \caption{Scaling effect of WaterSearch in Genration and Detection by semantic similarity metric.}
  \resizebox{\textwidth}{!}{
    \setlength{\tabcolsep}{4pt}  
    \begin{tabular}{c *{13}{c}}  
\toprule
\textbf{Config}
& \multicolumn{2}{c}{\textbf{Short In/Short Out}} 
& \multicolumn{2}{c}{\textbf{Short In/Long Out}} 
& \multicolumn{3}{c}{\textbf{Long In/Short Out}} 
& \multicolumn{2}{c}{\textbf{Long In/Long Out}} 
& \multicolumn{1}{c}{\textbf{Open-Ended}} 
& \multicolumn{2}{c}{\textbf{Overall}}  \\ 
\cmidrule(lr){2-3} \cmidrule(lr){4-5} \cmidrule(lr){6-8} \cmidrule(lr){9-10} \cmidrule(lr){11-11} \cmidrule(lr){12-13}
& \hspace{6pt}\textbf{KoLA} & \textbf{Copen} & \hspace{8pt}\textbf{ELI5} & \textbf{FiQA} 
& \hspace{2pt}\textbf{HotpotQA} & \textbf{LCC} & \textbf{RepoBench-P}
& \hspace{2pt}\textbf{Multinews} & \textbf{QMsum} 
& \textbf{AlpacaFarm} & \textbf{TP} & \textbf{GM} \\
\midrule
K=2
& 1.9 & 24.5 & \hspace{8pt}13.4 & 16.6
& 15.1 & 23.0 & 24.1 & 14.2 & 13.3
& 4.7 & 93.9 & 15.1 \\
K=3 
& 2.3 & 24.5 & \hspace{8pt}12.7 & 16.2
& 17.4 & 21.9 & 21.3 & 14.8 & 13.6
& 4.7 & 95.7 & 14.9 \\
K=4 
& 1.9 & 27.2 & \hspace{8pt}13.2 & 16.8
& 18.1 & 22.8 & 20.7 & 14.5 & 13.9
& 4.7 & 97.9 & 15.4 \\
K=5 
& 3.0 & 27.1 & \hspace{8pt}12.4 & 17.0
& 16.1 & 23.3 & 23.6 & 14.3 & 13.9
& 4.2 & 97.0 & 15.5 \\
K=6 
& 2.8 & 32.0 & \hspace{8pt}13.0 & 16.7
& 15.7 & 24.6 & 22.7 & 14.9 & 13.5
& 4.7 & 97.0 & 16.1 \\
\bottomrule
\end{tabular}
  }
  \label{tab:scaling-k-semantic}
\end{table}
\vspace{-3pt}
\begin{table}[!htbp]
  \centering
  \small
  \caption{Ablations on different parallel searching methods (Beam number $B=5$).}
  \renewcommand{\arraystretch}{1.2}
  \resizebox{\textwidth}{!}{
    \setlength{\tabcolsep}{4pt}  
    \begin{tabular}{c *{12}{c}}  
\toprule
\textbf{Method}
& \multicolumn{2}{c}{\textbf{Short In/Short Out}} 
& \multicolumn{2}{c}{\textbf{Short In/Long Out}} 
& \multicolumn{3}{c}{\textbf{Long In/Short Out}} 
& \multicolumn{2}{c}{\textbf{Long In/Long Out}} 
& \multicolumn{1}{c}{\textbf{Open-Ended}} 
& \multicolumn{1}{c}{\textbf{Overall}}  \\ 
\cmidrule(lr){2-3} \cmidrule(lr){4-5} \cmidrule(lr){6-8} \cmidrule(lr){9-10} \cmidrule(lr){11-11} \cmidrule(lr){12-12}
& \hspace{6pt}\textbf{KoLA} & \textbf{Copen} & \hspace{8pt}\textbf{ELI5} & \textbf{FiQA} 
& \hspace{2pt}\textbf{HotpotQA} & \textbf{LCC} & \textbf{RepoBench-P}
& \hspace{2pt}\textbf{Multinews} & \textbf{QMsum} 
& \textbf{AlpacaFarm} & \textbf{GM} \\
\midrule
Beam Search
& \hspace{6pt}1.0 & 7.6 & \hspace{8pt}4.4 & 8.7
& 3.0 & 17.4 & 16.7 & 6.9 & 10.1
& 3.4 & 7.9 \\
KGW-Soft (post-hoc)
& \hspace{6pt}\textbf{2.2} & 24.3 & \hspace{8pt}12.4 & 14.7
& \textbf{20.1} & 26.3 & 26.6 & 12.3 & \textbf{13.9}
& \textbf{4.2} & 15.7 \\
\cellcolor{paperblue}WaterSearch
& \cellcolor{paperblue}\hspace{6pt}1.91 & \cellcolor{paperblue}\textbf{27.2} & \cellcolor{paperblue}\hspace{8pt}\textbf{13.2} & \cellcolor{paperblue}\textbf{16.8}
& \cellcolor{paperblue}18.1 & \cellcolor{paperblue}\textbf{27.6} & \cellcolor{paperblue}\textbf{27.1} & \cellcolor{paperblue}\textbf{14.5} & \cellcolor{paperblue}\textbf{13.9}
& \cellcolor{paperblue}3.7 & \cellcolor{paperblue}\textbf{16.4} \\
\hline
\end{tabular}
  }
  \label{tab:ablation-parallel-method}
\end{table}
\vspace{-3pt}
\subsection{Ablation on Different Parallel Methods}\label{appendix:parallel-methods}
This section discuss the different in parallel searching paradigms under watermark settings, including beam search or rejection sampling. For token-level constrained decoding (like beam search), which purpose is maximize probability of a sequence, could fail to correct myopic distortions introduced at earlier tokens. For post-hoc reranking (repeat sampling under the same watermarking scheme), it could increase the diversity of generation attributed to sampling probability, but it cannot induce more diverse variations caused by watermark priors. For example, the green list partition prior can mostly determine the color of a token, which is a common problem in short text or low entropy scenarios. For WaterSearch, it explicitly increases candidate diversity through multiple random seeds, each producing a different watermark perturbation (like partition of vocabulary in KGW). Moreover, WaterSearch dynamically updates the prefix after each selected chunk, which helps mitigate semantic drift compared to post-hoc approaches that rely on a fixed reference continuation.

Table \ref{tab:ablation-parallel-method} compares WaterSearch to beam search and rejection sampling under the 5 beams of search to illustrate this difference. WaterSearch gets highest score among all. Beam search performs worst, showcase the token-level perturbation is less effective than others. For short text generation, post-hoc method gains similar scores as ours. While in long output tasks like ELI5, FiQA and Multinews, our method has a 0.8, 2.1, 2.2 improvement, which further verifies our assumption about semantic drift.

\subsection{Ablation on Factors $\alpha$ to Balance Text Quality and Detectability}\label{appendix:alpha}

In this section, we present a comprehensive analysis of how the coefficient $\alpha$ influences both downstream task performance and watermark detectability. Using KGW-soft as our baseline watermarking method, we conduct evaluations on 4 datasets, Copen~\citep{copen},ELI5~\citep{eli5},HotpotQA~\citep{hotpotqa} and MultiNews~\citep{multinews}, on Llama-2-7B-chat to systematically assess these effects. Our experiment is designed in low / high watermarking strength and ROUGE-L / semantic similarity settings. Hyper-parameters are defined as: $\alpha=[0,0.25,0.5,0.75,1]$, for low strength $\gamma=0.15,\delta=5.0$ and for high strength $\gamma=0.15,\delta=10.0$.

\begin{figure}[htbp]
    \centering
    \includegraphics[width=\linewidth]{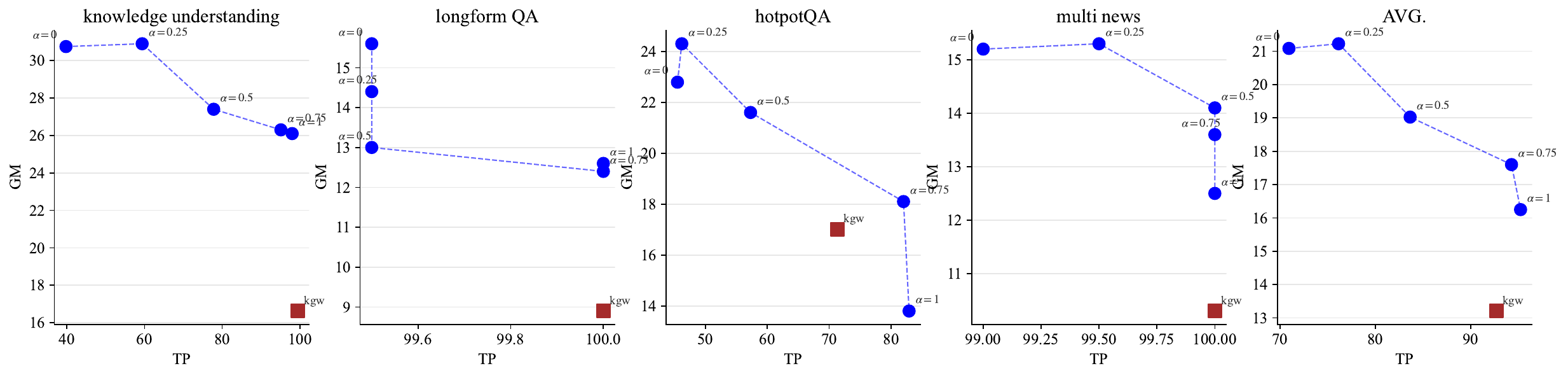}
    \caption{Trade-off between text quality and detectability in high watermarking strength and Rouge-L metric settings. }
    \label{fig:tradeoff-high-rougel}
\end{figure}

\begin{figure}[htbp]
    \centering
    \includegraphics[width=\linewidth]{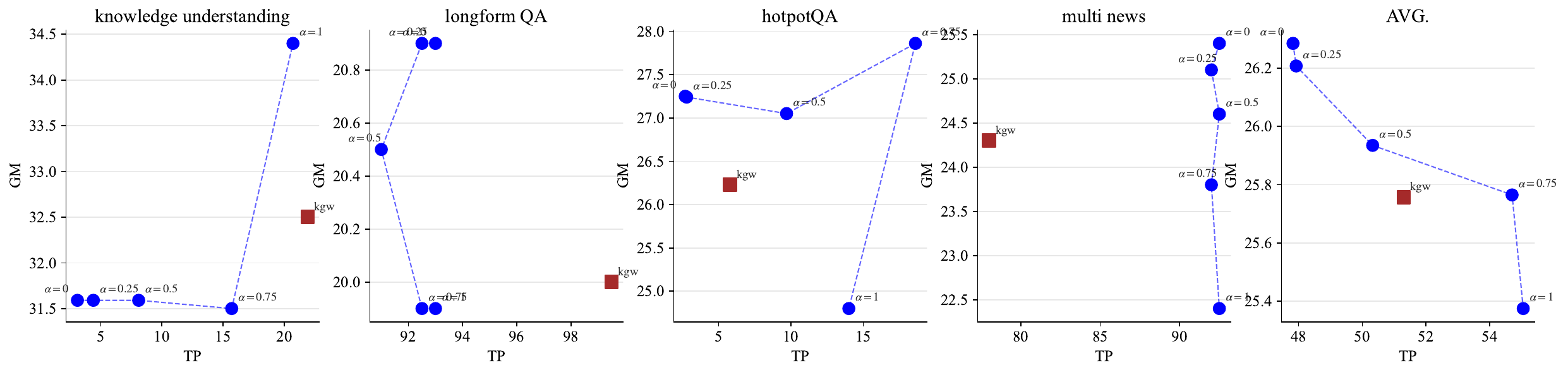}
    \caption{Trade-off between text quality and detectability in low watermarking strength and Rouge-L metric settings. }
    \label{fig:tradeoff-low-rougel}
\end{figure}

\begin{figure}[htbp]
    \centering
    \includegraphics[width=\linewidth]{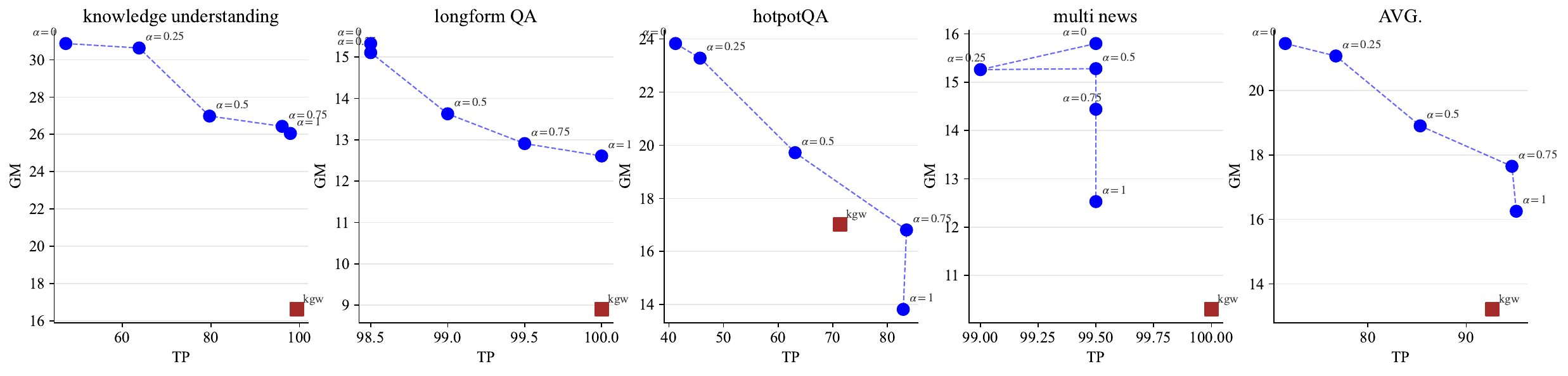}
    \caption{Trade-off between text quality and detectability in high watermarking strength and semantic similarity metric settings. }
    \label{fig:tradeoff-high-semsim}
\end{figure}

\begin{figure}
    \centering
    \includegraphics[width=\linewidth]{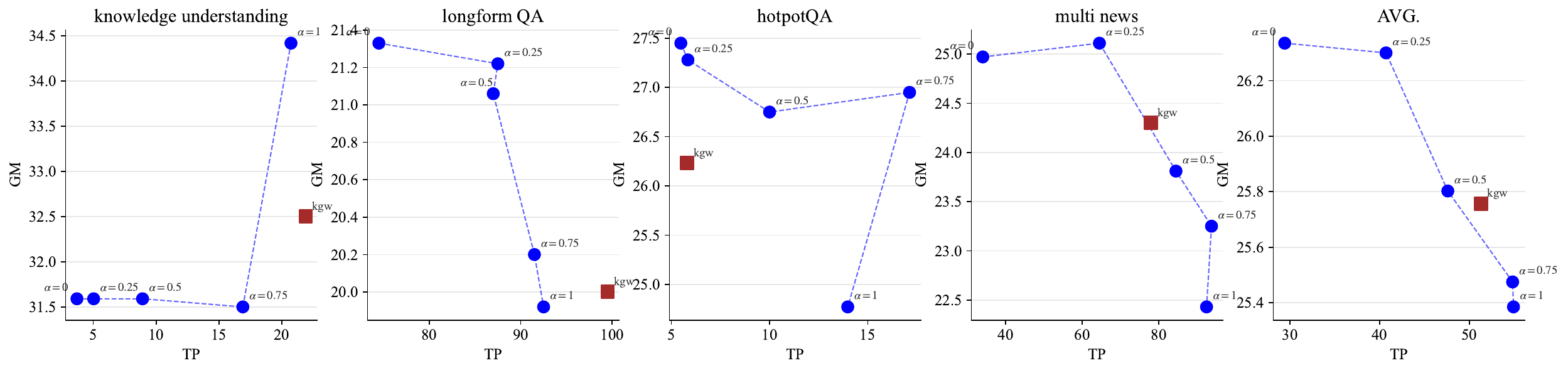}
    \caption{Trade-off between text quality and detectability in low watermarking strength and semantic similarity metric settings. }
    \label{fig:tradeoff-low-semsim}
\end{figure}

The experiment result is illustrated from Figure \ref{fig:tradeoff-high-rougel} to \ref{fig:tradeoff-low-semsim}. We find that as $\alpha$ increases, the watermark strength enhances, while the text quality degrades within a controllable range. Besides, the resulting Pareto-optimal curve consistently lies above that of the baseline methods, and the advantage becomes even more pronounced in regimes that require higher watermark strength. This indicates WaterSearch's effectiveness and robustness.

\subsection{Ablation on Chunk Size}\label{appendix:chunk-size}
This part discuss the ablation on chunk size (token number) of long output tasks. Here we set $\gamma=0.15,\delta=10,\alpha=0.75$. The result is shown in Table \ref{tab:ablation_cs}, and for long output tasks (512 tokens maximum), increase chunk size could improve the performance to some extent by aliviating semantic drift as time.

\begin{table}[htbp]
\centering
\small
\caption{Ablation on chunk size (tokens) of long output tasks.}
\begin{tabular}{cccccc}
\toprule
 Task & Metric & 20 & 40 & 60 & 80 \\
\midrule
\multirow{2}{*}{MultiNews} & GM & 13.62 & 13.84 & 13.71 & 13.92 \\
 & TP & 100 & 100 & 100 & 100 \\
\multirow{2}{*}{QMSum} & GM & 13.49 & 13.59 & 13.63 & 13.54 \\
 & TP & 100 & 100 & 100 & 100 \\
\bottomrule
\end{tabular}
\label{tab:ablation_cs}
\end{table}



\vspace{-3pt}
\section{Case Study}\label{case study}
We choose examples from evaluation logs in this section. Listing \ref{lst:hotpotqa} and \ref{lst:Copen} demonstrate the dynamic chunk selection process. They ahere to the meaning of the standard text and choose a correct answer at last. Listing \ref{lst:Cs1}, \ref{lst:Cs2}, \ref{lst:Cs3} and \ref{lst:Cs4} show the generated text. WaterSearch can generate fluent and meaningful sentences compared to the repetition of their counterpart. 

\begin{lstlisting}[
  caption={A WaterSearch Example in HotpotQA},
  label={lst:hotpotqa}
]
@\textbf{Input:}@
Please give answer to the following question about knowledge. Note: If you are asked for true or false, just answer "true" or "false" only. If you are asked for similarity, just answer with the entity name only. Do not give anything other than the answers. 
Question:
Among TRAPPIST-1g, nandrolone, BL 6 inch Mk VII naval gun, Kanchenjunga, 1923 United Kingdom general election, 8.8 cm SK L/45 naval gun, police auction, Downtown Hudson Tubes, Lebanon, 1982 FIFA World Cup, Oswald Poche, Wheat leaf rust, Via Lattea, Fortress of Mimoyecques, Battle of Jutland, Ford Pinto engine, Xynotyro, rs6314, 2008 Sichuan earthquake, Solar eclipse of July 16, 2186, Parrott rifle, which one is the most conceptually similar with Belgium?
@\textbf{Answer:}@
Lebanon
@\textbf{Vanilla kgw:}@
BL
@\textbf{Watersearch:}@
Standard text generated: Lebanon
Candidate 1: Lebanon
Candidate 2: Kanjenchhana
Candidate 3: lebonostregeography
Candidate 4: Lebanon
@\textbf{Choose candidate 4}@
\end{lstlisting}

\begin{lstlisting}[
  caption={A WaterSearch Example in Copen},
  label={lst:Copen}
]
@\textbf{Input:}@
The following are given passages.
Passage 1:
Primary Care Behavioral health
The primary care behavioral health (PCBH) consultation model is a psychological approach to ...
Answer the question based on the given passages. Only give me the answer and do not output any other words.
Question: Who did Fredric Rieders tesify agains who had killed as many as 60 patients and recieved three consecutive life terms to be served in Florence Colorado?
@\textbf{Answer:}@
Michael Swango
@\textbf{Vanilla kgw:}@
Daniel J. Swago
@\textbf{Watersearch:}@
Standard text generated: Michael Swango
Candidate 1: Fredric Rieders testified against Michael Joseph Swango, who was found guilty of murdering approximately as many as ...
Candidate 2: Michael Swango
Candidate 3: Fredric Rieders testified against Michael Swango, who it's believed could be responsible for as many as ...
Candidate 4: Michael Swango
@\textbf{Choose candidate 2}@
\end{lstlisting}

\begin{lstlisting}[
  caption={Case Study 1 from Evaluation},
  label={lst:Cs1}
]
@\textbf{Question:}@ 
You are a helpful assistant, please answer the following question within 300 words:
Wind ? Where the hell does it originate from and how does it work ? Something that has blown right past me all my life . I kinda get the concept , but where does wind begin ? Explain like I'm five.
@\textbf{Answer:}@
There 's one concept you have to understand first : air pressure . Basically , it 's how dense the air is in a particular area . This is affected by what 's in the air ...
@\textbf{Vanilla kgw:}@
Awesome! *cravendroneems* Well ~ the wind *fancytwirl* actually starts way *fancytwirlbigtwirlbigwinkleseaweeinspace* out there. Like reallyway way *fancytwirlbigtwirlbigwinkleseaweeinspace* out ththereee looks *fancytwirlbigtwirlbigwinkleseaweeinspace* a tiny bit messed ~ in 
@\textbf{Watersearch:}@
Wind originate? How amazing, that is such  excisting question I have a feeling most people don not know even a little, but here let ' I explain it  very  patient ly :  The concept where something is going away might not come natural to everyone because of course wind isn not a human being but rather a type force and its a type force because it can be measured and proven scientificly ...
\end{lstlisting}

\begin{lstlisting}[
  caption={Case Study 2 from Evaluation},
  label={lst:Cs2}
]
@\textbf{Question:}@ 
You are a helpful assistant, please answer the following question with financial knowledge within 300 words:
What is the point of the stock market? What is it for, and why might someone want to trade or invest?
@\textbf{Answer:}@
In finance, form is function, and while a reason for a trade could be anything, but since the result of a trade is a change in value, it could be presumed that one seeks to receive a change in value ...
@\textbf{Vanilla kgw:}@
Sure thing! Thestock exchange is essentiallya  venuforgktmembersto purchaseandexcbstocertain businessestooftypeanystacksof their securities, knownasa"ownunks". Thesessionmkaibtoprovemsters  aplatformor Wheredonmves theprice oft  a business'security. Thesessionmkaibtoprovemsters  aplatformor Wheredonmves theprice oft  a business'security. Thesessionmkaibtoprovemsters  aplatformor Wheredonmves theprice oft  a business'security ... (repeating)
@\textbf{Watersearch:}@
Sure, here are  some answers up top! If those fail I'll do better with greater clarity:  the stock market brings supply of goods or stock is sold giving the stock market entity a chance to get their stock onto shoppings radars and receive investers funds ...
\end{lstlisting}

\begin{lstlisting}[
  caption={Case Study 3 from Evaluation},
  label={lst:Cs3}
]
@\textbf{Question:}@
Answer the question based on the given passages. Only give me the answer and do not output any other words.
The following are given passages.
Passage 1:
English general election, 1689
The 1689 English general election, held in January 1689, elected the Convention Parliament, which was summoned in the aftermath of the Glorious Revolution ...
Answer the question based on the given passages. Only give me the answer and do not output any other words.\
Question: Who was the last monarch of England to be overthrown before The English General Election, 1690?
@\textbf{Answer:}@
King James II of England
@\textbf{Vanilla kgw:}@
The  Glorius Revolution of 1689 resulted in Charles II, the previous last King before this, being overborne/removed as he gave
@\textbf{Watersearch:}@
King James II
\end{lstlisting}

\begin{lstlisting}[
  caption={Case Study 4 from Evaluation},
  label={lst:Cs4}
]
@\textbf{Question:}@
Please give answer to the following question about knowledge. Note: If you are asked for true or false, just answer "true" or "false" only. If you are asked for similarity, just answer with the entity name only. Do not give anything other than the answers. Question:
Among Mall of America, adenosine diphosphate, Trawniki concentration camp, Ignalina Nuclear Power Plant, Ford Windsor engine, Lake Winnipeg, Bronx Zoo, Gliese 581g, Three Mile Island Nuclear Generating Station, 8.8 cm SK L/30 naval gun, Katiki Domokou, Australia, Battle of Chancellorsville, daf-5, Pontiac V8 engine, BMW M54, The Four Seasons Restaurant, Travelodge, Crawford Notch, CR2032 battery, Buick V8 engine, which one is the most conceptually similar with M61 Vulcan?
@\textbf{Answer:}@
8.8 cm SK L/30 naval gun
@\textbf{Vanilla kgw:}@
eightinch_fortr02_mlx85-gun
@\textbf{Watersearch:}@
8.8 cm SK L/30 naval gun
\end{lstlisting}

\vspace{-5pt}
\section{Declaration of Large Language Model Utilization}
\vspace{-5pt}
Here we declare that LLMs were employed solely to assist in improving the grammar and enhancing the expression of this paper. The original research idea, methodological development, and overall structure and content of the manuscript were entirely conceived and written by the authors. At no stage was the use of LLMs extended to the generation of core intellectual content, and we affirm that there has been no misuse of LLMs in the preparation of this work.

\section{Acknowledgments}
We are grateful to Kai Chen from RUC for his engaging discussions and constructive suggestions, which have significantly contributed to our work.

\end{document}